\documentclass[letterpaper]{article} 
\usepackage[]{aaai2026}  
\usepackage{times}  
\usepackage{helvet}  
\usepackage{courier}  
\usepackage[hyphens]{url}  
\usepackage{graphicx} 
\usepackage{natbib}  
\usepackage{caption} 
\usepackage{algorithm}
\usepackage{algorithmic}
\usepackage[utf8]{inputenc} 
\usepackage{latexsym}
\usepackage{microtype}
\usepackage{nicefrac}
\usepackage{amsmath,amsfonts,amssymb,amsthm}
\usepackage[table]{xcolor}
\usepackage{booktabs}
\usepackage{multirow}
\usepackage{array,tabularx,makecell}
\usepackage{adjustbox}
\usepackage{subcaption}
\usepackage{placeins}
\usepackage[most]{tcolorbox}
\usepackage{enumitem}

\theoremstyle{plain}
\newtheorem{theorem}{Theorem}
\newtheorem{lemma}[theorem]{Lemma}

\theoremstyle{definition}

\allowdisplaybreaks[1]
\setlist{nosep,leftmargin=*}

\definecolor{headercolor}{RGB}{220,228,240}
\definecolor{lgtlight}{RGB}{225,238,228}
\newcolumntype{L}[1]{>{\raggedright\arraybackslash}p{#1}}
\newcolumntype{C}[1]{>{\centering\arraybackslash}p{#1}}
\newcolumntype{Y}{>{\raggedright\arraybackslash}X}

\newcommand{\hdr}{\rowcolor{headercolor}}
\newcommand{\lgtrow}{\rowcolor{lgtlight}}
\newcommand{\tightrules}{\setlength{\aboverulesep}{0pt}\setlength{\belowrulesep}{0pt}}
\definecolor{tbdred}{RGB}{200,30,30}

\newcommand{\hstrutA}{\rule{0pt}{2.9ex}}
\newcommand{\hstrutB}{\rule[-1.0ex]{0pt}{1.2ex}}

\usepackage{newfloat}
\usepackage{listings}
\DeclareCaptionStyle{ruled}{labelfont=normalfont,labelsep=colon,strut=off} 
\floatstyle{ruled}
\newfloat{listing}{tb}{lst}{}
\floatname{listing}{Listing}
\title{Language-Guided Tuning: Configuration Optimization for Automated ML Research}
\author{
    Yuxing Lu\textsuperscript{\rm 1}\equalcontrib\thanks{Corresponding Author (yxlu0613@gmail.com).},
    Yucheng Hu\textsuperscript{\rm 2}\equalcontrib,
    Nan Sun\textsuperscript{\rm 3},
    Xukai Zhao\textsuperscript{\rm 4}
}
\affiliations{
    \textsuperscript{\rm 1}Peking University\\
    \textsuperscript{\rm 2}Southeast University\\
    \textsuperscript{\rm 3}Huazhong University of Science and Technology\\
    \textsuperscript{\rm 4}Tsinghua University\\

}

\begin{document}
\nocopyright
\maketitle

\begin{abstract}
Configuration optimization remains a critical bottleneck in machine learning, requiring coordinated tuning across model architecture, training strategy, feature engineering, and hyperparameters. Traditional approaches treat these dimensions independently and lack interpretability, while recent automated methods struggle with dynamic adaptability and semantic reasoning about optimization decisions. We introduce Language-Guided Tuning (LGT), a framework that employs multi-agent Large Language Models to automatically optimize configurations through natural language reasoning. We apply textual feedback signals that complement numerical optimization by providing semantic understanding of training dynamics and configuration interdependencies. LGT coordinates three specialized agents: an Advisor that proposes configuration changes, an Evaluator that assesses progress, and an Optimizer that refines the decision-making process, creating a self-improving feedback loop. Through comprehensive evaluation on seven diverse datasets, LGT demonstrates substantial improvements over traditional optimization methods while maintaining high interpretability.
\end{abstract}

\section{Introduction}
\label{sec:intro}
The success of deep learning has been accompanied by an increasingly complex challenge: the optimization of neural network performance through configuration tuning across multiple interdependent components~\citep{huang2019survey,alom2019state}. Modern deep learning systems require optimization across four critical dimensions: model architecture, feature engineering, training strategy, and hyperparameters~\citep{deng2014tutorial,pouyanfar2018survey,goodfellow2016deep}. The process of identifying optimal configurations across this multi-dimensional space has emerged as a critical bottleneck in the machine learning development pipeline, often requiring extensive computational resources and domain expertise to navigate the complex interactions between these components~\citep{yu2020hyper}. For instance, the choice of data augmentation strategy may require corresponding adjustments to learning rate and model architecture~\citep{shorten2019survey}, while the selection of loss functions can impact the effectiveness of different optimization algorithms~\citep{wang2022comprehensive}.

Current approaches to configuration optimization fall into several categories, each with significant limitations that hinder their effectiveness in addressing the multi-dimensional nature of the problem~\citep{huang2019survey}. Manual tuning is inherently unscalable and heavily dependent on practitioner expertise, especially when coordinating across multiple configuration dimensions. Grid search and random search methods suffer from the curse of dimensionality and fail to leverage information from previous evaluations, requiring exhaustive exploration of the configuration space~\citep{liashchynskyi2019grid}. More sophisticated approaches such as Bayesian optimization algorithms offer improvements in efficiency but remain computationally expensive and struggle with the dynamic nature of training processes~\citep{victoria2021automatic}. Some of these methods do adapt during training: Population Based Training, which we evaluate as a baseline, mutates hyperparameters online from performance feedback, but the adaptation is driven by scalar scores alone, so it cannot use the semantics of a training trace, and the configuration dimensions are still treated as independent variables rather than as an interdependent whole. Additionally, the decision-making process in these approaches lacks interpretability and reasoning capabilities, making it difficult to understand and trust the optimization choices made by the system.

Recent advances in large language models (LLMs) have demonstrated remarkable capabilities for optimization tasks, with applications in hyperparameter tuning and configuration selection~\citep{liu2025agenthpo,chen2022towards,yang2023large}. However, existing LLM-based optimization approaches face critical limitations: they typically operate as single-agent systems without multi-dimensional reasoning, lack feedback mechanisms for iterative improvement, and fail to provide interpretable optimization trajectories that combine semantic understanding with numerical precision.

\begin{figure*}[t]
    \centering
    \includegraphics[width=0.8\textwidth]{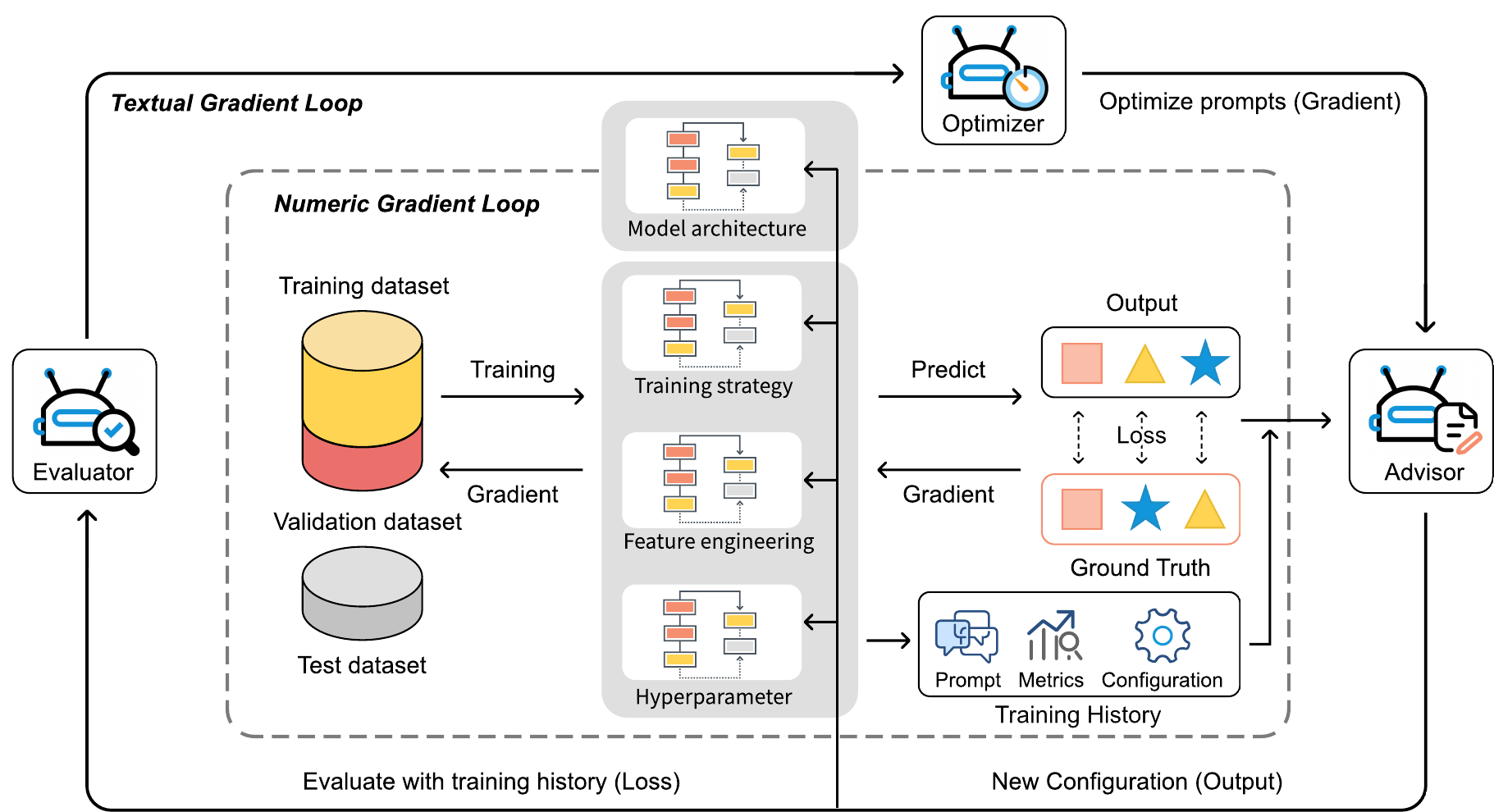}
    \caption{\textbf{Overall framework of Language-Guided Tuning (LGT).} The system operates through dual loops: an inner numeric gradient loop for traditional training, and an outer textual gradient loop where three LLM agents (Advisor, Evaluator, Optimizer) coordinate to optimize configurations across four dimensions (model architecture, training strategy, feature engineering, hyperparameters).}
    \label{fig:lgt_framework}
    \vspace{-10pt}
\end{figure*}

To address these limitations, we propose Language-Guided Tuning (LGT), a multi-agent framework that leverages the strength of LLMs and their textual feedback signals that complement numerical loss functions through natural language reasoning. Our system employs three specialized LLM agents: an \textbf{Advisor} that proposes configuration changes, an \textbf{Evaluator} that assesses progress, and an \textbf{Optimizer} that refines prompts for continuous improvement, creating a self-improving feedback loop for coordinated multi-dimensional optimization. Experimental validation on seven diverse datasets shows that LGT matches or exceeds all 11 baselines on every dataset, with the largest margins on the harder tabular and generation tasks, while exposing a natural-language rationale for every accepted edit.

Language-mediated optimization is not itself new, and we do not claim it. Textual
feedback as a first-class optimization signal was established by
TextGrad~\citep{yuksekgonul2024textgrad} and GEPA~\citep{agrawal2025gepa};
role-specialized agents driven by a critic text appear in
AutoDAN-Turbo~\citep{liu2025autodanturbo}; LLMs as hyperparameter proposers
appear in AgentHPO~\citep{liu2025agenthpo} and
OptiMindTune~\citep{madiraju2025optimindtune}; and agents that edit training code,
launch runs and keep or discard them by score are already practical in
autoresearch~\citep{karpathy2025autoresearch} and The AI
Scientist~\citep{lu2024ai}. What distinguishes LGT is not the use of
language but \emph{where in the training pipeline the language acts}. Our
contributions are:

\noindent\textbf{(1) The edit unit is an epoch boundary, not a run.} Automated
research agents treat a full training run as the atom of action: propose, train
to completion, score, keep or discard. LGT instead edits a run that is already in
flight, which changes the cost regime rather than the interface: the entire
outer loop costs $1.06\times$ plain training (Table~\ref{tab:gen_robust_eff}c),
against the $N\times$ implied by run-level search.

\noindent\textbf{(2) A validation-gated acceptance rule that makes online editing
safe.} Editing a live run is only useful if a bad LLM proposal cannot wreck it. A
candidate is trained one epoch from the incumbent checkpoint and promoted only if
it clears the incumbent by a margin $\eta_t$ on a held-out split; otherwise the
checkpoint is restored (Algorithm~\ref{alg:lgt}). This makes the run monotone by
construction and is what lets us state a convergence result
(Appendix~\ref{sec:Theoretical Proofs}): none of the LLM-based tuning systems
above offer one.

\noindent\textbf{(3) Joint coordination across four axes inside one live model.}
Architecture, feature engineering, training strategy and hyperparameters are
edited against a shared state, so a proposal must remain consistent with weights
and optimizer state that already exist. Prompt-space methods such as
GEPA do not face this constraint; single-axis tuners such as AgentHPO and DARTS
do not face the coupling.

We therefore position LGT not as a competitor to autoresearch-style systems but as the inner loop such a system currently lacks: the component an outer research agent can call when it needs a configuration improved cheaply, safely and with an auditable reason attached.

\section{Related Works}

\subsection{Configuration Tuning Methods}
Grid and Random Search~\citep{liashchynskyi2019grid,bergstra2012random} scale poorly and ignore structure. Bayesian and SMBO toolkits add surrogates-canonical BO~\citep{frazier2018tutorial}, Optuna's TPE~\citep{akiba2019optunanextgenerationhyperparameteroptimization}, and SMAC3's random forests~\citep{lindauer2022smac3}, while multi-fidelity schedulers~\citep{li2018hyperband,falkner2018bohb} reallocate budget; both remain score-driven without semantic reasoning. NAS methods such as DARTS~\citep{liu2018darts} target architecture alone, and AutoML pipelines~\citep{karmaker2021automl,zoller2021benchmark} decouple design axes and justify changes in human-readable form. LGT instead proposes semantically justified edits that reference training traces, coordinates architecture, augmentation, loss, optimizer, scheduler, and hyperparameters in a single outer loop, and applies a validation-gated accept/reject rule online at epoch boundaries.

\subsection{Automated Research Agents}
A parallel line of work puts LLM agents in charge of the whole experimental
loop. autoresearch~\citep{karpathy2025autoresearch} and The AI
Scientist~\citep{lu2024ai} let an agent edit training code, launch a
run, read the result and keep or revert the change; MLE-bench~\citep{ICLR2025_7e3767db}
measures such agents on Kaggle-style engineering tasks. These systems and LGT
differ in granularity rather than in ambition. Their atom of action is a
completed run, which is what makes them general: an agent that can rewrite code
can change anything, but also what makes them expensive, since every candidate
costs a full training budget. LGT deliberately gives up that
generality: it edits a fixed configuration schema at epoch boundaries inside one
run. We read the
two as composable, an outer research agent proposing what to try, an inner loop
like LGT settling how to configure it, rather than as alternatives.

\subsection{LLM-Based Optimization}
LLMs can act as planners and critics that propose and justify edits in natural language~\citep{zhao2023survey,wang2025history,yu2024deep,zhang2023using}, and language-mediated feedback has been shown to guide iterative improvement across domains~\citep{yuksekgonul2024textgrad}. Applied to configuration tuning, LLMs leverage priors over architectures, optimization, and regularization to suggest changes conditioned on training traces~\citep{liu2025agenthpo,yang2023large}. Closely related multi-agent systems, including OptiMindTune and MAT-Agent, further decompose tuning into specialized agents for HPO or adaptive training control~\citep{madiraju2025optimindtune,zhangmat}.Yet existing LLM-for-HPO systems remain largely heuristic ~\citep{liu2025agenthpo,huang2025calm,nie2024importance,ma2024llamoco}, and struggle under nonstationary dynamics or delayed improvements.  LGT closes this gap with multi-agent reasoning under a validation-gated acceptance rule: Advisor, Evaluator, and Prompt-Optimizer agents produce edits grounded in training signals, and proposals are accepted only when they pass a statistical improvement margin, coordinating all six design axes online while keeping updates interpretable.

\section{Methods}

\subsection{Problem Formulation}

Figure~\ref{fig:lgt_framework} presents the overall architecture of LGT, illustrating the integration of textual and numerical gradient flows for comprehensive configuration optimization.

\subsubsection{Configuration Optimization}

We formulate the configuration optimization problem as follows. Let $\mathcal{C} = \mathcal{C}_A \times \mathcal{C}_F \times \mathcal{C}_T \times \mathcal{C}_H$ be the multi-dimensional configuration space, where $\mathcal{C}_A$, $\mathcal{C}_F$, $\mathcal{C}_T$, and $\mathcal{C}_H$ represent the spaces of model architecture, feature engineering, training strategy, and hyperparameters respectively. We adopt a hierarchical optimization schedule. The model architecture, given its large impact, is optimized only once after a full training run. In contrast, feature engineering, training strategy, and hyperparameters are optimized sequentially within each epoch, and each subsequent optimization step has full access to the solutions chosen in previous steps. Each configuration $c \in \mathcal{C}$ parametrizes a model $f_c$ that maps inputs $x \in \mathcal{X}$ to outputs $y \in \mathcal{Y}$. The objective is to find the optimal configuration $c^* \in \mathcal{C}$ that minimizes the expected loss over the data distribution:
\begin{equation}
\footnotesize
c^* = \arg\min_{c \in \mathcal{C}} \mathcal{L}(c) = \arg\min_{c \in \mathcal{C}} \mathbb{E}_{(x,y) \sim \mathcal{D}}[\ell(f_c(x), y)]
\end{equation}
where $\ell$ is the task-specific loss function and $\mathcal{D}$ is the data distribution. The key challenge lies in the complexity and interdependencies between configurations, making $\mathcal{C}$ non-separable and requiring optimization across all dimensions.

\subsubsection{Multi-Agent System}

Our framework employs a multi-agent system $\mathcal{A} = \{\mathbb{A}_{\text{adv}}, \mathbb{A}_{\text{eval}}, \mathbb{A}_{\text{opt}}\}$ consisting of three agents: an Advisor ($\mathbb{A}_{\text{adv}}$) that generates configuration modifications, an Evaluator ($\mathbb{A}_{\text{eval}}$) that assesses optimization effectiveness, and a Prompt Optimizer ($\mathbb{A}_{\text{opt}}$) that refines agent decision-making capabilities.

The three agents have distinct codomains, and we keep them separate rather than
folding them into one transition function. Writing $\mathcal{M}$ for the space
of training metrics, $\mathcal{H}$ for optimization histories, $\mathcal{P}$ for
prompts and $\Sigma^{*}$ for natural-language strings, the agents are
$\mathbb{A}_{\mathrm{adv}}: \mathcal{C} \times \mathcal{M} \times \mathcal{H} \times \mathcal{P} \rightarrow \Sigma^{*}$,
$\mathbb{A}_{\mathrm{eval}}: (\mathcal{C} \times \mathcal{M})^{2} \times \mathcal{H} \rightarrow \Sigma^{*}$
and
$\mathbb{A}_{\mathrm{opt}}: \mathcal{M} \times \mathcal{C} \times \mathcal{P} \times \Sigma^{*} \rightarrow \Sigma^{*}$.
Each returns free text; dedicated parsers
$\mathcal{P}_{\text{parse}}^{\text{cfg}}, \mathcal{P}_{\text{parse}}^{\text{dec}}, \mathcal{P}_{\text{parse}}^{\text{sgg}}, \mathcal{P}_{\text{parse}}^{\text{prm}}$
map that text into a candidate configuration, a binary accept/reject decision, a
suggestion, and a revised prompt respectively. Only
$\mathcal{P}_{\text{parse}}^{\text{cfg}} \circ \mathbb{A}_{\mathrm{adv}}$ ever
produces an element of $\mathcal{C}$, so the configuration space is modified
exclusively along that path. This creates a closed-loop system where agents communicate through structured pipelines, enabling coordinated optimization while maintaining interpretability through natural language reasoning. We provide the prompt templates for each agent in Appendix~\ref{sec:prompts}. In Appendix~\ref{sec:agent_outputs}, we present the outputs from agent interactions, including cases of successes and failures.

\subsection{Configuration Components}
\subsubsection{Model Architecture Optimization}
The architecture space $\mathcal{C}_A = \{\theta_a | \theta_a \in \mathcal{F}_a\}$ encompasses all neural network configurations, including layer types, connectivity patterns, and structural parameters, where $\mathcal{F}_a$ defines the space of architectures satisfying computational and structural constraints. The architecture optimization function : $\mathcal{M}_t \times \mathcal{C}_A \rightarrow \mathcal{C}_A$ analyzes current training metrics $\mathcal{M}_t$ to generate improved architectures. The optimal architecture selection follows:
\begin{equation}
\footnotesize
\theta_a^* = \arg\min_{\theta_a \in \mathcal{C}_A} \mathbb{E}[\mathcal{L}(f_{\theta_a}(x), y)]
\end{equation}
where $f_{\theta_a}$ denotes the model with architecture $\theta_a$. This optimization occurs independently before training cycles of other configurations (Appendix~\ref{Model Architecture Agent}).

\subsubsection{Feature Engineering}
The feature engineering space $\mathcal{C}_F = \{\tau: \mathcal{X} \rightarrow \mathcal{X}' | \tau \in \mathcal{T}\}$ consists of all possible data transformation functions, where $\mathcal{T}$ represents available methods including geometric transformations, noise injection, and domain-specific operations. The augmentation selection function : $\mathcal{M}_t \times \mathcal{C}_F \rightarrow \mathcal{C}_F$ adapts strategies based on current training metrics $\mathcal{M}_t$. The augmentation selection follows:
\begin{equation}
\footnotesize
\tau^* = \arg\min_{\tau \in \mathcal{C}_F} \mathbb{E}[\mathcal{L}(f(\tau(x)), y)]
\end{equation}
where $f$ is the current model. This enables dynamic response to fitting patterns through appropriate augmentation method selection (Appendix~\ref{Feature Engineering Agent}).

\subsubsection{Training Strategy Optimization}
The training strategy space $\mathcal{C}_T = \mathcal{L}_{\text{loss}} \times \mathcal{O}_{\text{opt}} \times \mathcal{S}_{\text{sched}}$ encompasses loss functions, optimizers, and learning rate schedulers. The strategy selection function : $\mathcal{M}_t \times \mathcal{C}_T \rightarrow \mathcal{C}_T$ adapts training configurations based on current training metrics $\mathcal{M}_t$. The strategy selection follows:
\begin{equation}
\footnotesize
(l^*, o^*, s^*) = \arg\min_{(l,o,s) \in \mathcal{C}_T} \mathbb{E}[\mathcal{L}(f_{l,o,s}(x), y)]
\end{equation}
where $f_{l,o,s}$ denotes loss function, optimizer, and scheduler respectively (Appendix~\ref{Training Strategy Agent}).

\subsubsection{Hyperparameter Optimization}
The hyperparameter space $\mathcal{C}_H = \{\lambda | \lambda_i \in [\lambda_i^{\min}, \lambda_i^{\max}]\}$ includes learning rates, weight decay, and other parameters. The hyperparameter update function : $\mathcal{M}_t \times \mathcal{C}_H \rightarrow \mathcal{C}_H$ modifies parameters based on training metrics $\mathcal{M}_t$. The epoch-level update:
\begin{equation}
\footnotesize
\lambda_{t+1} = \Pi_{\mathcal{C}_H}[\lambda_t + \Delta\lambda_t]
\end{equation}
where $\Delta\lambda_t$ minimizes the expected loss $\mathbb{E}[\mathcal{L}(f_{\lambda}(x), y)]$, and $\Pi_{\mathcal{C}_H}$ ensures constraint satisfaction (Appendix~\ref{Hyperparameter Optimization Agent}).

\subsection{Multi-Agents and Prompt Optimization}

\subsubsection{Advisor (Output): config generator}

The Advisor ($\mathbb{A}_{\text{adv}}$) analyzes the current training state and generates new configuration recommendations. The advisor function $\mathbb{A}_{\text{adv}}: (c_t, \mathcal{M}_t, \mathcal{C}) \rightarrow \Delta c_t^{\text{NL}}$ maps the current configuration, training metrics, and configuration space to natural language configuration recommendations $\Delta c_t^{\text{NL}}$:
\begin{equation}
\footnotesize
\Delta c_t^{\text{NL}} = \mathcal{G}_{\text{LLM}}(\text{prompt}_{\text{adv}}(c_t, \mathcal{M}_t, \mathcal{C}))
\end{equation}
The new configuration $c_{t+1}$ is obtained through parsing:
\begin{equation}
\footnotesize
c_{t+1} = \mathcal{P}_{\text{parse}}^{\text{cfg}}(\Delta c_t^{\text{NL}})
\end{equation}
where $\mathcal{P}_{\text{parse}}^{\text{cfg}}$ extracts and instantiates modified configuration from the natural language outputs.

\subsubsection{Evaluator (Loss): performance assessor}

The Evaluator ($\mathbb{A}_{\text{eval}}$) judges the proposed change with both a binary acceptance signal and a suggestion for improvement. The mapping $\mathbb{A}_{\text{eval}}:\ (c_t,\ \mathcal{M}_t,\ c_{t-1},\ \mathcal{M}_{t-1})\ \rightarrow\ \mathcal{I}_{\text{eval}}^{\text{NL}}$
is given by
\begin{equation}
\footnotesize
\mathcal{I}_{\text{eval}}^{\text{NL}}=\mathcal{G}_{\text{LLM}}\!\left(\text{prompt}_{\text{eval}}(c_t,\mathcal{M}_t,\ c_{t-1},\mathcal{M}_{t-1})\right)
\end{equation}
Parsing yields a decision and a suggestion:
\footnotesize
\begin{align}
\text{success}_t &= \mathcal{P}_{\text{parse}}^{\text{dec}}\!\left(\mathcal{I}_{\text{eval}}^{\text{NL}}\right)\in\{0,1\},\\
s_t^{\text{NL}} &= \mathcal{P}_{\text{parse}}^{\text{sgg}}\!\left(\mathcal{I}_{\text{eval}}^{\text{NL}}\right),
\end{align}
\normalsize
where $s_t^{\text{NL}}$ is a natural-language suggestion that may reference specific configuration fields, search constraints, or prompt refinements (Appendix~\ref{subsec:evaluator_prompts}).

\subsubsection{Optimizer (Gradient): prompt updater}
The Prompt Optimizer ($\mathbb{A}_{\text{opt}}$) refines the Advisor prompt using current performance, Advisor prompt, and Evaluator’s suggestion. The mapping $\mathbb{A}_{\text{opt}}:\ (\mathcal{H}_t,\ \mathcal{M}_t,\ c_t,\ \text{prompt}_{\text{adv}}^{t},\ s_t^{\text{NL}})\ \rightarrow\ \Delta p_t^{\text{NL}}$:
\begin{equation}
\footnotesize
\Delta p_t^{\text{NL}}=\mathcal{G}_{\text{LLM}}\!\left(\text{prompt}_{\text{opt}}(\mathcal{M}_t,c_t,\ \text{prompt}_{\text{adv}}^{t},\ s_t^{\text{NL}})\right)
\end{equation}
The updated advisor prompt is obtained through:
\begin{equation}
\footnotesize
\text{prompt}_{\text{adv}}^{t+1}=\mathcal{P}_{\text{parse}}^{\text{prm}}(\Delta p_t^{\text{NL}})
\end{equation}
where $\mathcal{P}_{\text{parse}}^\text{prm}$ extracts and applies the prompt modifications from the outputs (Appendix~\ref{subsec:optimizer_prompts}).

\subsubsection{Self-Improving Feedback Loop}

The collaboration creates a self-improving, meta-level feedback loop that operates across all configuration dimensions. The system maintains an optimization history $\mathcal{H}_t$ that accumulates comprehensive information from each epoch, providing the necessary context for leveraging inter-dependencies between configuration components that traditional methods treat independently:
\begin{equation}
\footnotesize
\mathcal{H}_t = \mathcal{H}_{t-1} \cup \{(c_t, \mathcal{M}_t, \Delta c_t^{\text{NL}},\ \mathcal{I}_{\text{eval}}^{\text{NL}}, \text{prompt}_t)\}
\end{equation}

The feedback loop among Advisor, Evaluator and Optimizer ensures the training process becomes adaptive and targeted by enabling agents to refer to both training results and configurations for making appropriate adjustments.

\section{Experiments}
This section mainly introduces the experimental details. To validate generalization, we provide supplementary experiments using a complex ViT architecture in Appendix~\ref{sec:appendix-experiments}.
\subsection{Datasets}
\begin{table}[t]
\centering
\caption{Dataset characteristics.}
\vspace{-6pt}
\label{tab:datasets}
\footnotesize
\renewcommand{\arraystretch}{0.7}
\setlength{\tabcolsep}{3.2pt}
\begin{tabular}{lrrlr}
\toprule
\hdr
\textbf{Dataset} & \textbf{Samples} & \textbf{Features} & \textbf{Task Type} & \textbf{Classes} \\
\midrule
MNIST            & 70,000 & 784   & Classification & 10 \\
CIFAR-10         & 60,000 & 3,072 & Classification & 10 \\
Iris             & 150    & 4     & Classification & 3  \\
Water Potability & 3,276  & 9     & Classification & 2  \\
House Price      & 50,000 & 5     & Regression     & -- \\
Wine Quality     & 1,599  & 11    & Regression     & -- \\
Tiny Shakespeare & 40,000 & --    & Generation     & -- \\
\bottomrule
\end{tabular}
\vspace{-15pt}
\end{table}
We evaluate LGT on seven datasets spanning three fundamental machine learning tasks (Table~\ref{tab:datasets}, Appendix~\ref{sec:dataset}). For image classification, we use MNIST~\citep{deng2012mnist} and CIFAR-10~\citep{krizhevsky2009cifar} to assess performance on visual recognition tasks. For tabular data, we include classification tasks (Iris~\citep{goodfellow2016deep}, Water Potability~\citep{simeonov2001environmetric}) and regression tasks (House Price~\citep{harrison1978hedonic}, Wine Quality~\citep{cortez2009modeling}) to demonstrate effectiveness on structured data. For text generation, we employ Tiny Shakespeare~\citep{karpathy2015charrnn} to evaluate sequence modeling capabilities. This comprehensive evaluation demonstrates the framework's versatility across different data modalities and task types.



\begin{table*}[t]
\centering
\caption{Classification performance comparison.}
\vspace{-5pt}
\label{tab:overall_results}
\footnotesize
\tightrules
\setlength{\tabcolsep}{3.4pt}
\renewcommand{\arraystretch}{0.75}
\begin{adjustbox}{max width=\linewidth}
\begin{tabular}{lccc ccc ccc ccc}
\toprule
\hdr
 & \multicolumn{3}{c}{\textbf{MNIST}\hstrutA} & \multicolumn{3}{c}{\textbf{CIFAR-10}} & \multicolumn{3}{c}{\textbf{Water Potability}} & \multicolumn{3}{c}{\textbf{Iris}} \\
\cmidrule(lr){2-4}\cmidrule(lr){5-7}\cmidrule(lr){8-10}\cmidrule(l){11-13}
\hdr
\textbf{Method} & \textbf{Acc.} & \textbf{F1} & \textbf{AUC} & \textbf{Acc.} & \textbf{F1} & \textbf{AUC} & \textbf{Acc.} & \textbf{F1} & \textbf{AUC} & \textbf{Acc.} & \textbf{F1} & \textbf{AUC}\hstrutB \\
\midrule
\addlinespace[2pt]
No Tuning & 96.77$_{\text{\scriptsize 0.2}}$ & 96.78$_{\text{\scriptsize 0.2}}$ & 98.37$_{\text{\scriptsize 0.08}}$ & 69.01$_{\text{\scriptsize 2.4}}$ & 69.01$_{\text{\scriptsize 2.5}}$ & 93.60$_{\text{\scriptsize 1.2}}$ & 65.17$_{\text{\scriptsize 2.8}}$ & 50.94$_{\text{\scriptsize 3.1}}$ & 65.33$_{\text{\scriptsize 1.9}}$ & 93.33$_{\text{\scriptsize 4.2}}$ & 92.95$_{\text{\scriptsize 3.8}}$ & 96.60$_{\text{\scriptsize 2.1}}$ \\
Random Search & 97.87$_{\text{\scriptsize 0.1}}$ & 97.87$_{\text{\scriptsize 0.1}}$ & 99.96$_{\text{\scriptsize 0.04}}$ & 74.57$_{\text{\scriptsize 4.6}}$ & 74.48$_{\text{\scriptsize 4.5}}$ & 96.95$_{\text{\scriptsize 0.8}}$ & 65.15$_{\text{\scriptsize 2.0}}$ & 52.11$_{\text{\scriptsize 3.6}}$ & 66.05$_{\text{\scriptsize 2.5}}$ & 94.33$_{\text{\scriptsize 3.7}}$ & 94.45$_{\text{\scriptsize 3.2}}$ & 96.78$_{\text{\scriptsize 1.8}}$ \\
Grid Search & 97.75$_{\text{\scriptsize 0.1}}$ & 97.75$_{\text{\scriptsize 0.1}}$ & 99.96$_{\text{\scriptsize 0.05}}$ & 76.82$_{\text{\scriptsize 1.4}}$ & 76.70$_{\text{\scriptsize 1.4}}$ & 97.19$_{\text{\scriptsize 0.2}}$ & 65.89$_{\text{\scriptsize 2.9}}$ & 61.67$_{\text{\scriptsize 3.2}}$ & 66.78$_{\text{\scriptsize 2.0}}$ & 95.67$_{\text{\scriptsize 2.1}}$ & 94.23$_{\text{\scriptsize 2.9}}$ & 97.45$_{\text{\scriptsize 2.0}}$ \\
Bayesian Opt. & 97.67$_{\text{\scriptsize 0.3}}$ & 97.66$_{\text{\scriptsize 0.3}}$ & 98.45$_{\text{\scriptsize 0.20}}$ & 72.45$_{\text{\scriptsize 2.1}}$ & 72.23$_{\text{\scriptsize 2.2}}$ & 94.95$_{\text{\scriptsize 0.9}}$ & 66.67$_{\text{\scriptsize 2.7}}$ & 62.45$_{\text{\scriptsize 2.9}}$ & 67.23$_{\text{\scriptsize 1.8}}$ & 95.67$_{\text{\scriptsize 1.2}}$ & 94.45$_{\text{\scriptsize 1.7}}$ & 97.78$_{\text{\scriptsize 1.5}}$ \\
Optuna & 97.79$_{\text{\scriptsize 0.1}}$ & 97.79$_{\text{\scriptsize 0.1}}$ & 99.96$_{\text{\scriptsize 0.04}}$ & 76.83$_{\text{\scriptsize 4.9}}$ & 76.80$_{\text{\scriptsize 0.8}}$ & 97.24$_{\text{\scriptsize 0.1}}$ & 66.50$_{\text{\scriptsize 1.4}}$ & 45.42$_{\text{\scriptsize 3.0}}$ & 68.39$_{\text{\scriptsize 2.2}}$ & 90.33$_{\text{\scriptsize 4.6}}$ & 90.00$_{\text{\scriptsize 5.1}}$ & 98.40$_{\text{\scriptsize 1.6}}$ \\
SMAC3 & 97.61$_{\text{\scriptsize 0.2}}$ & 97.61$_{\text{\scriptsize 0.2}}$ & 99.95$_{\text{\scriptsize 0.05}}$ & 74.97$_{\text{\scriptsize 2.3}}$ & 75.02$_{\text{\scriptsize 2.3}}$ & 96.81$_{\text{\scriptsize 0.5}}$ & 66.16$_{\text{\scriptsize 1.1}}$ & 47.96$_{\text{\scriptsize 3.2}}$ & 68.92$_{\text{\scriptsize 2.1}}$ & 94.67$_{\text{\scriptsize 3.7}}$ & 94.62$_{\text{\scriptsize 3.8}}$ & 99.53$_{\text{\scriptsize 0.5}}$ \\
KerasTuner & 97.78$_{\text{\scriptsize 0.1}}$ & 97.78$_{\text{\scriptsize 0.1}}$ & 99.96$_{\text{\scriptsize 0.04}}$ & 77.69$_{\text{\scriptsize 1.3}}$ & 77.59$_{\text{\scriptsize 1.4}}$ & 97.34$_{\text{\scriptsize 0.2}}$ & 63.05$_{\text{\scriptsize 1.4}}$ & 50.77$_{\text{\scriptsize 3.8}}$ & 66.06$_{\text{\scriptsize 2.1}}$ & 95.33$_{\text{\scriptsize 4.0}}$ & 95.34$_{\text{\scriptsize 4.0}}$ & 98.67$_{\text{\scriptsize 1.7}}$ \\
Hyperband & 97.32$_{\text{\scriptsize 0.2}}$ & 97.31$_{\text{\scriptsize 0.2}}$ & 99.93$_{\text{\scriptsize 0.06}}$ & 75.14$_{\text{\scriptsize 3.4}}$ & 74.97$_{\text{\scriptsize 3.3}}$ & 97.04$_{\text{\scriptsize 0.6}}$ & 63.89$_{\text{\scriptsize 1.9}}$ & 51.03$_{\text{\scriptsize 3.2}}$ & 65.48$_{\text{\scriptsize 1.7}}$ & 95.00$_{\text{\scriptsize 3.1}}$ & 94.99$_{\text{\scriptsize 3.1}}$ & 99.33$_{\text{\scriptsize 0.9}}$ \\
BOHB & 97.47$_{\text{\scriptsize 0.3}}$ & 97.46$_{\text{\scriptsize 0.3}}$ & 99.94$_{\text{\scriptsize 0.05}}$ & 77.42$_{\text{\scriptsize 0.9}}$ & 77.34$_{\text{\scriptsize 1.0}}$ & 97.43$_{\text{\scriptsize 0.1}}$ & 66.13$_{\text{\scriptsize 1.9}}$ & 50.98$_{\text{\scriptsize 2.6}}$ & 68.52$_{\text{\scriptsize 2.0}}$ & 95.00$_{\text{\scriptsize 4.0}}$ & 94.96$_{\text{\scriptsize 4.1}}$ & 99.30$_{\text{\scriptsize 0.7}}$ \\
PBT & 97.05$_{\text{\scriptsize 0.2}}$ & 97.05$_{\text{\scriptsize 0.2}}$ & 99.92$_{\text{\scriptsize 0.07}}$ & 77.11$_{\text{\scriptsize 2.3}}$ & 76.95$_{\text{\scriptsize 2.3}}$ & 97.40$_{\text{\scriptsize 0.3}}$ & 66.78$_{\text{\scriptsize 1.3}}$ & 46.72$_{\text{\scriptsize 2.9}}$ & 69.05$_{\text{\scriptsize 2.2}}$ & 95.67$_{\text{\scriptsize 3.3}}$ & 95.65$_{\text{\scriptsize 3.4}}$ & 99.27$_{\text{\scriptsize 0.7}}$ \\
DARTS & 97.34$_{\text{\scriptsize 0.5}}$ & 97.34$_{\text{\scriptsize 0.5}}$ & 98.67$_{\text{\scriptsize 0.21}}$ & 75.89$_{\text{\scriptsize 3.4}}$ & 75.67$_{\text{\scriptsize 3.5}}$ & 96.45$_{\text{\scriptsize 1.3}}$ & 66.12$_{\text{\scriptsize 3.1}}$ & 61.89$_{\text{\scriptsize 3.3}}$ & 66.89$_{\text{\scriptsize 2.2}}$ & 94.33$_{\text{\scriptsize 2.8}}$ & 94.67$_{\text{\scriptsize 2.1}}$ & 98.12$_{\text{\scriptsize 1.2}}$ \\
\lgtrow\textbf{LGT} & \textbf{98.99}$_{\text{\scriptsize 0.8}}$ & \textbf{98.99}$_{\text{\scriptsize 0.8}}$ & \textbf{99.99}$_{\text{\scriptsize 0.10}}$ & \textbf{79.64}$_{\text{\scriptsize 2.0}}$ & \textbf{79.61}$_{\text{\scriptsize 2.1}}$ & \textbf{98.42}$_{\text{\scriptsize 0.7}}$ & \textbf{66.92}$_{\text{\scriptsize 2.5}}$ & \textbf{63.52}$_{\text{\scriptsize 2.7}}$ & \textbf{69.60}$_{\text{\scriptsize 1.6}}$ & \textbf{96.67}$_{\text{\scriptsize 1.8}}$ & \textbf{96.67}$_{\text{\scriptsize 1.7}}$ & \textbf{99.54}$_{\text{\scriptsize 0.8}}$ \\
\bottomrule
\end{tabular}
\end{adjustbox}
\end{table*}

\begin{table*}[t]
\centering
\caption{Regression and text generation performance comparison.}
\vspace{-5pt}
\label{tab:regression_generation_results}
\label{tab:regression_results}
\label{tab:shakespeare_results}
\footnotesize
\tightrules
\setlength{\tabcolsep}{6.8pt}
\renewcommand{\arraystretch}{0.75}
\begin{adjustbox}{max width=\linewidth}
\begin{tabular}{lccc ccc ccc}
\toprule
\hdr
 & \multicolumn{3}{c}{\textbf{House Price}\hstrutA} & \multicolumn{3}{c}{\textbf{Wine Quality}} & \multicolumn{3}{c}{\textbf{Tiny Shakespeare}} \\
\cmidrule(lr){2-4}\cmidrule(lr){5-7}\cmidrule(l){8-10}
\hdr
\textbf{Method} & \textbf{MAE} & \textbf{MSE} & \textbf{R$^2$} & \textbf{MAE} & \textbf{MSE} & \textbf{R$^2$} & \textbf{PPL} & \textbf{BLEU-4} & \textbf{Acc.}\hstrutB \\
\midrule
\addlinespace[2pt]
No Tuning & 0.458$_{\text{\scriptsize 0.026}}$ & 0.329$_{\text{\scriptsize 0.030}}$ & 0.425$_{\text{\scriptsize 0.03}}$ & 0.706$_{\text{\scriptsize 0.032}}$ & 0.828$_{\text{\scriptsize 0.060}}$ & -0.267$_{\text{\scriptsize 0.05}}$ & 2.96$_{\text{\scriptsize 0.12}}$ & 0.681$_{\text{\scriptsize 0.023}}$ & 82.45$_{\text{\scriptsize 1.8}}$ \\
Random Search & 0.441$_{\text{\scriptsize 0.029}}$ & 0.313$_{\text{\scriptsize 0.031}}$ & 0.451$_{\text{\scriptsize 0.03}}$ & 0.680$_{\text{\scriptsize 0.030}}$ & 0.781$_{\text{\scriptsize 0.055}}$ & -0.210$_{\text{\scriptsize 0.04}}$ & 2.43$_{\text{\scriptsize 0.09}}$ & 0.724$_{\text{\scriptsize 0.018}}$ & 84.12$_{\text{\scriptsize 1.5}}$ \\
Grid Search & 0.435$_{\text{\scriptsize 0.027}}$ & 0.305$_{\text{\scriptsize 0.029}}$ & 0.463$_{\text{\scriptsize 0.03}}$ & 0.665$_{\text{\scriptsize 0.030}}$ & 0.746$_{\text{\scriptsize 0.052}}$ & -0.160$_{\text{\scriptsize 0.05}}$ & 2.38$_{\text{\scriptsize 0.08}}$ & 0.731$_{\text{\scriptsize 0.016}}$ & 84.56$_{\text{\scriptsize 1.3}}$ \\
Bayesian Opt. & 0.420$_{\text{\scriptsize 0.026}}$ & 0.289$_{\text{\scriptsize 0.027}}$ & 0.505$_{\text{\scriptsize 0.02}}$ & 0.632$_{\text{\scriptsize 0.028}}$ & 0.691$_{\text{\scriptsize 0.048}}$ & -0.090$_{\text{\scriptsize 0.04}}$ & 2.51$_{\text{\scriptsize 0.11}}$ & 0.712$_{\text{\scriptsize 0.020}}$ & 83.89$_{\text{\scriptsize 1.6}}$ \\
Optuna & 0.418$_{\text{\scriptsize 0.025}}$ & 0.286$_{\text{\scriptsize 0.026}}$ & 0.511$_{\text{\scriptsize 0.02}}$ & 0.629$_{\text{\scriptsize 0.027}}$ & 0.686$_{\text{\scriptsize 0.047}}$ & -0.080$_{\text{\scriptsize 0.04}}$ & 2.35$_{\text{\scriptsize 0.07}}$ & 0.738$_{\text{\scriptsize 0.015}}$ & 84.78$_{\text{\scriptsize 1.2}}$ \\
SMAC3 & 0.419$_{\text{\scriptsize 0.026}}$ & 0.287$_{\text{\scriptsize 0.026}}$ & 0.508$_{\text{\scriptsize 0.02}}$ & 0.630$_{\text{\scriptsize 0.028}}$ & 0.689$_{\text{\scriptsize 0.047}}$ & -0.075$_{\text{\scriptsize 0.01}}$ & 2.47$_{\text{\scriptsize 0.10}}$ & 0.719$_{\text{\scriptsize 0.019}}$ & 84.23$_{\text{\scriptsize 1.4}}$ \\
KerasTuner & 0.430$_{\text{\scriptsize 0.028}}$ & 0.298$_{\text{\scriptsize 0.028}}$ & 0.489$_{\text{\scriptsize 0.02}}$ & 0.642$_{\text{\scriptsize 0.029}}$ & 0.711$_{\text{\scriptsize 0.050}}$ & -0.120$_{\text{\scriptsize 0.04}}$ & 2.33$_{\text{\scriptsize 0.06}}$ & 0.742$_{\text{\scriptsize 0.014}}$ & 84.91$_{\text{\scriptsize 1.1}}$ \\
Hyperband & 0.432$_{\text{\scriptsize 0.028}}$ & 0.300$_{\text{\scriptsize 0.029}}$ & 0.485$_{\text{\scriptsize 0.03}}$ & 0.653$_{\text{\scriptsize 0.030}}$ & 0.731$_{\text{\scriptsize 0.052}}$ & -0.150$_{\text{\scriptsize 0.07}}$ & 2.54$_{\text{\scriptsize 0.13}}$ & 0.708$_{\text{\scriptsize 0.021}}$ & 83.67$_{\text{\scriptsize 1.7}}$ \\
BOHB & 0.417$_{\text{\scriptsize 0.024}}$ & 0.284$_{\text{\scriptsize 0.025}}$ & 0.515$_{\text{\scriptsize 0.02}}$ & 0.617$_{\text{\scriptsize 0.026}}$ & 0.668$_{\text{\scriptsize 0.045}}$ & 0.010$_{\text{\scriptsize 0.03}}$ & 2.41$_{\text{\scriptsize 0.08}}$ & 0.728$_{\text{\scriptsize 0.017}}$ & 84.34$_{\text{\scriptsize 1.3}}$ \\
PBT & 0.414$_{\text{\scriptsize 0.023}}$ & 0.279$_{\text{\scriptsize 0.024}}$ & 0.525$_{\text{\scriptsize 0.02}}$ & 0.609$_{\text{\scriptsize 0.025}}$ & 0.655$_{\text{\scriptsize 0.042}}$ & 0.035$_{\text{\scriptsize 0.06}}$ & 2.29$_{\text{\scriptsize 0.05}}$ & 0.749$_{\text{\scriptsize 0.013}}$ & 85.12$_{\text{\scriptsize 1.0}}$ \\
DART/DARTS & 0.408$_{\text{\scriptsize 0.022}}$ & 0.262$_{\text{\scriptsize 0.025}}$ & 0.539$_{\text{\scriptsize 0.02}}$ & 0.588$_{\text{\scriptsize 0.024}}$ & 0.613$_{\text{\scriptsize 0.040}}$ & 0.081$_{\text{\scriptsize 0.03}}$ & 2.62$_{\text{\scriptsize 0.14}}$ & 0.695$_{\text{\scriptsize 0.022}}$ & 83.21$_{\text{\scriptsize 1.9}}$ \\
\lgtrow\textbf{LGT} & \textbf{0.402}$_{\text{\scriptsize 0.021}}$ & \textbf{0.254}$_{\text{\scriptsize 0.024}}$ & \textbf{0.557}$_{\text{\scriptsize 0.02}}$ & \textbf{0.547}$_{\text{\scriptsize 0.027}}$ & \textbf{0.515}$_{\text{\scriptsize 0.045}}$ & \textbf{0.212}$_{\text{\scriptsize 0.03}}$ & \textbf{1.72}$_{\text{\scriptsize 0.04}}$ & \textbf{0.907}$_{\text{\scriptsize 0.008}}$ & \textbf{92.50}$_{\text{\scriptsize 0.7}}$ \\
\bottomrule
\end{tabular}
\end{adjustbox}
\vspace{-5pt}
\end{table*}

\subsection{Experimental Setup}

\textbf{LLM Backbone:} All LLM agents use DeepSeek-V3.2~\citep{liu2024deepseek} as backbone with temperature set as 0.2.

\noindent\textbf{Splits and what the agents may see:} Every dataset is split 80/20 into train and test; the validation signal that gates configuration updates is a further 10\% carved out of the \emph{training} portion, so the final split is 72/8/20. The Evaluator, the Advisor and the accept/reject rule read only the validation split; the test split is touched exactly once, after the last epoch, to produce the numbers we report. This matters because the framework would otherwise be optimising directly against the test set. Confirm against the data-loading code before submission and state the exact percentages.

\noindent\textbf{Search Spaces:} Our optimization framework explores multiple dimensions simultaneously. The architecture search examines networks with 2--5 hidden layers with 32--512 neurons. Feature engineering encompasses geometric transformations (rotation, scaling), data augmentation (noise injection), and task-specific preprocessing strategies. We evaluate different optimization algorithms like Adam, SGD, RMSprop, AdamW, Adagrad, and AdaFactor. Additional hyperparameters include class weights ranging from 0.1 to 10.0, and learning rates spanning from $10^{-4}$ to $10^{-1}$.

\noindent\textbf{Evaluation:} We adopt an evaluation protocol where each optimization method explores up to 50 configurations and trained for 10 epochs to balance computational efficiency and performance assessment. To ensure statistical validity, all experiments are repeated across 10 independent runs using random seeds 42--51. Performance metrics are task-specific: classification tasks employ accuracy, F1-score, and AUC; regression tasks utilize MAE, MSE, and R$^2$; while text generation tasks use perplexity, BLEU, and token-level accuracy.

\section{Results}

\subsection{Overall Performance Comparison}

Tables~\ref{tab:overall_results} and~\ref{tab:regression_generation_results} present the comprehensive performance comparison between LGT and established optimization methods across all datasets. Given that conventional optimization methods cannot easily coordinate multi-dimensional search, and that LLM-based tuning approaches capable of such coordination remain rare and largely closed-source, we necessarily compare against these established traditional baselines. The experimental results are scaled to ensure four significant figures and comparability across the same metric. Results show mean±standard deviation over \textbf{10} independent runs, demonstrating LGT's superiority with low variance.

Across classification, regression, and generation, LGT delivers consistent gains with low variance (10 runs; mean$\pm$sd). On classification tasks, LGT improves accuracy by up to +10.63 points over no-tuning baselines. LGT also surpasses the strongest baselines (e.g., CIFAR-10 vs.\ KerasTuner, +1.95\% on accuracy). On regression and NLP tasks, LGT also achieves better performance. Unlike black-box optimizers that provide limited visibility into parameter updates, LGT exposes textual rationales for each configuration change. These textual feedback trace \emph{why} a change is proposed, \emph{what} is modified, and \emph{how} it affects metrics, yielding transparent and auditable optimization trajectories (Appendix~\ref{subsec:success_example}).

The results indicate a shift from search-driven tuning to semantics-guided optimization. LGT leverages natural-language reasoning to coordinate coupled design model architecture, training strategy, feature engineering, and hyperparameters. This enables multi-dimensional, interdependent updates that other methods typically treat as independent, delivering stronger performance across different tasks while preserving interpretability.

\begin{figure}[t]
\centering
\includegraphics[width=\linewidth]{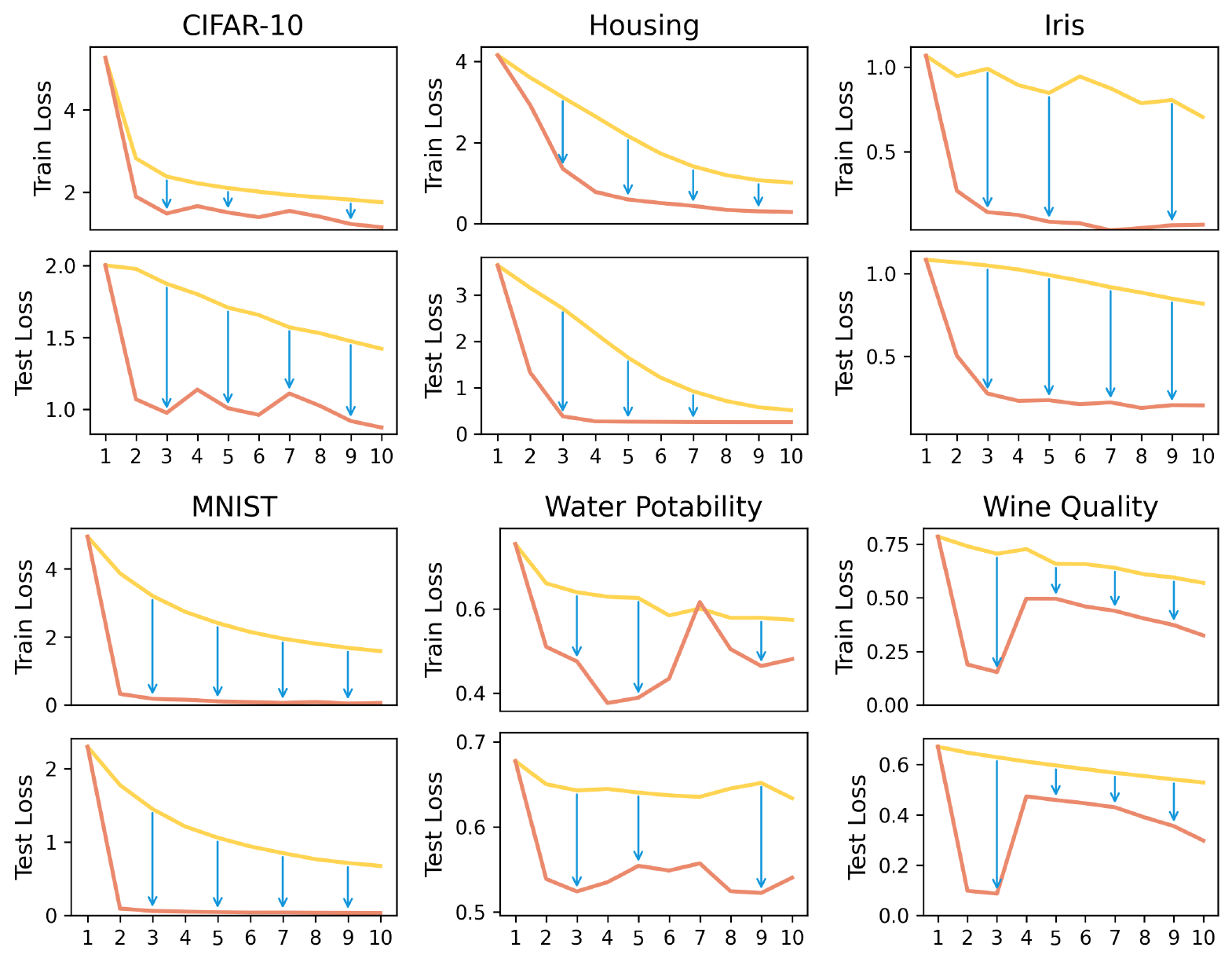}
\caption{Loss curves across 6 datasets comparing LGT with baseline. LGT demonstrates faster convergence and better performance.}
\label{fig:convergence_curves}
\vspace{-10pt}
\end{figure}

\subsection{Dynamic Convergence Analysis}

Figure~\ref{fig:convergence_curves} illustrates the training dynamics of LGT compared to No Tuning baseline across 6 datasets. LGT achieves faster convergence and superior final performance compared to the baseline, with improvements visible across all datasets. The performance gaps are substantial on MNIST (final loss: 0.0469 vs 1.5827 baseline) and Iris (final loss: 0.0337 vs 0.7041 baseline), demonstrating LGT's ability to achieve better optimization outcomes.

The convergence analysis reveals why LGT outperforms traditional optimization approaches. Unlike methods that follow predetermined schedules or rely on fixed heuristics, LGT evaluates training dynamics and implements intelligent configuration adjustments throughout the training process without requiring additional computational overhead. This dynamic adaptation enables faster convergence compared to static baselines, with the red curves descending more than orange baseline curves across all datasets. The stability advantage becomes evident when comparing LGT's trajectories to the oscillatory behavior of grid search or random search methods. While traditional approaches often struggle with premature convergence or unstable training dynamics, LGT maintains steady progress toward optimal configurations through its semantic understanding of training feedback.

\begin{figure}[t]
\centering
\includegraphics[width=\linewidth]{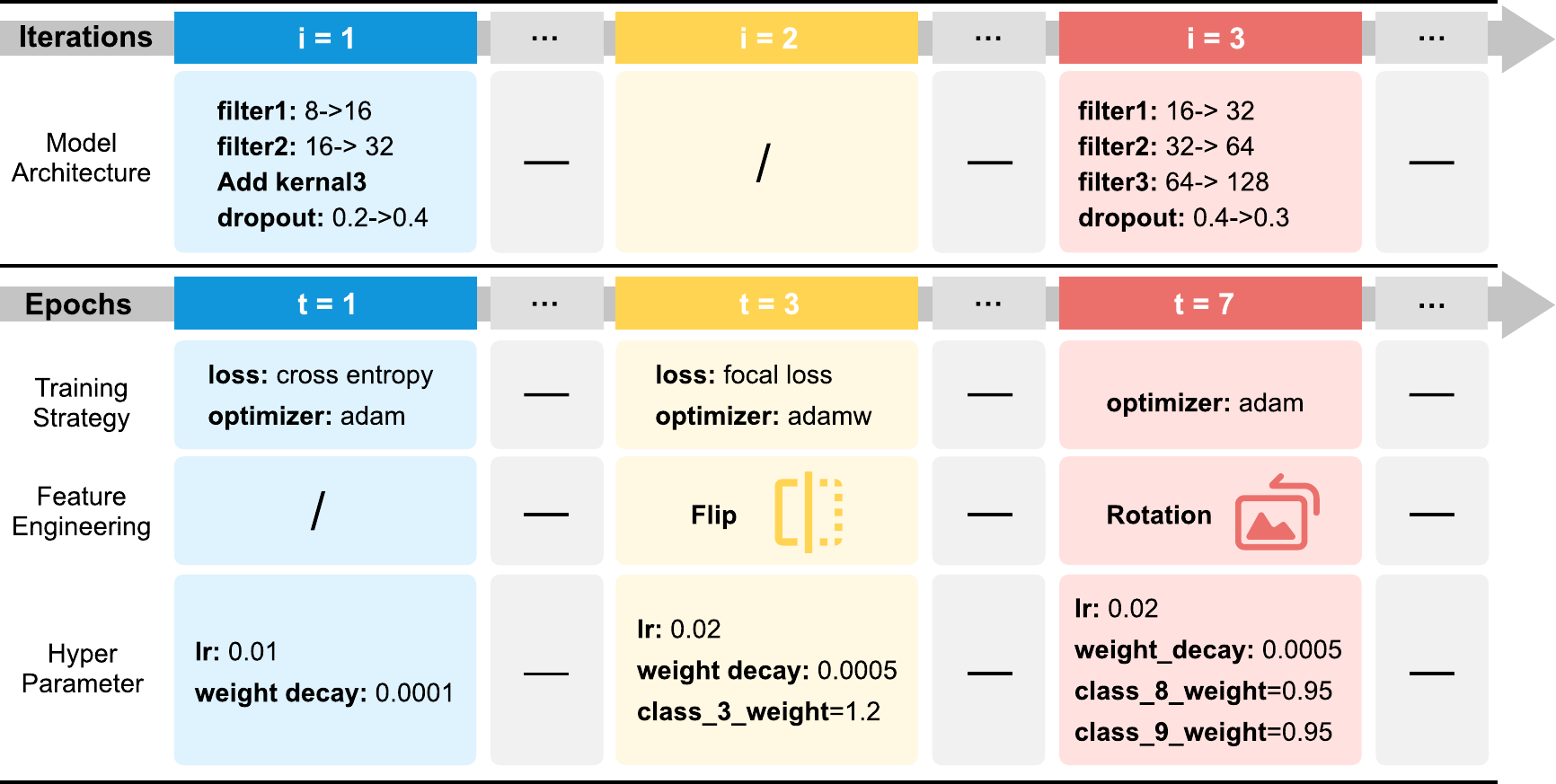}
\caption{Configuration evolution showing coordinated optimization across four dimensions.}
\label{fig:config_evolution}
\vspace{-10pt}
\end{figure}

Figure~\ref{fig:config_evolution} demonstrates the evolution of configuration dimensions across both strategic iterations and tactical epoch-level adjustments. The dual timeline reveals LGT's intelligent coordination strategy: iteration-level planning establishes major architectural foundations, while epoch-level adaptation fine-tunes training dynamics in real-time. At the iteration level, LGT follows a structured progression: Iteration 1 focuses on architectural modifications (filter1: 8→16, filter2: 16→32, adding kernel3, dropout: 0.2→0.4), while Iteration 3 scales up the architecture further with refined regularization (dropout: 0.4→0.3). the epoch-level timeline shows tactical adjustments within training cycles: at epoch t=1, the system begins with cross-entropy loss and Adam optimizer, transitions to focal loss and AdamW at epoch t=3 to address emerging class imbalance issues, then optimizes back to Adam at epoch t=7 for final convergence acceleration. The feature engineering progression from basic augmentation (flip) to transformations (rotation) and the hyperparameter refinements (learning rates from 0.01→0.02→0.02) demonstrate how LGT coordinates configuration without creating optimization conflicts.

\subsection{Textual Evolution and Interpretability}

\begin{figure}[t]
\centering
\includegraphics[width=\linewidth]{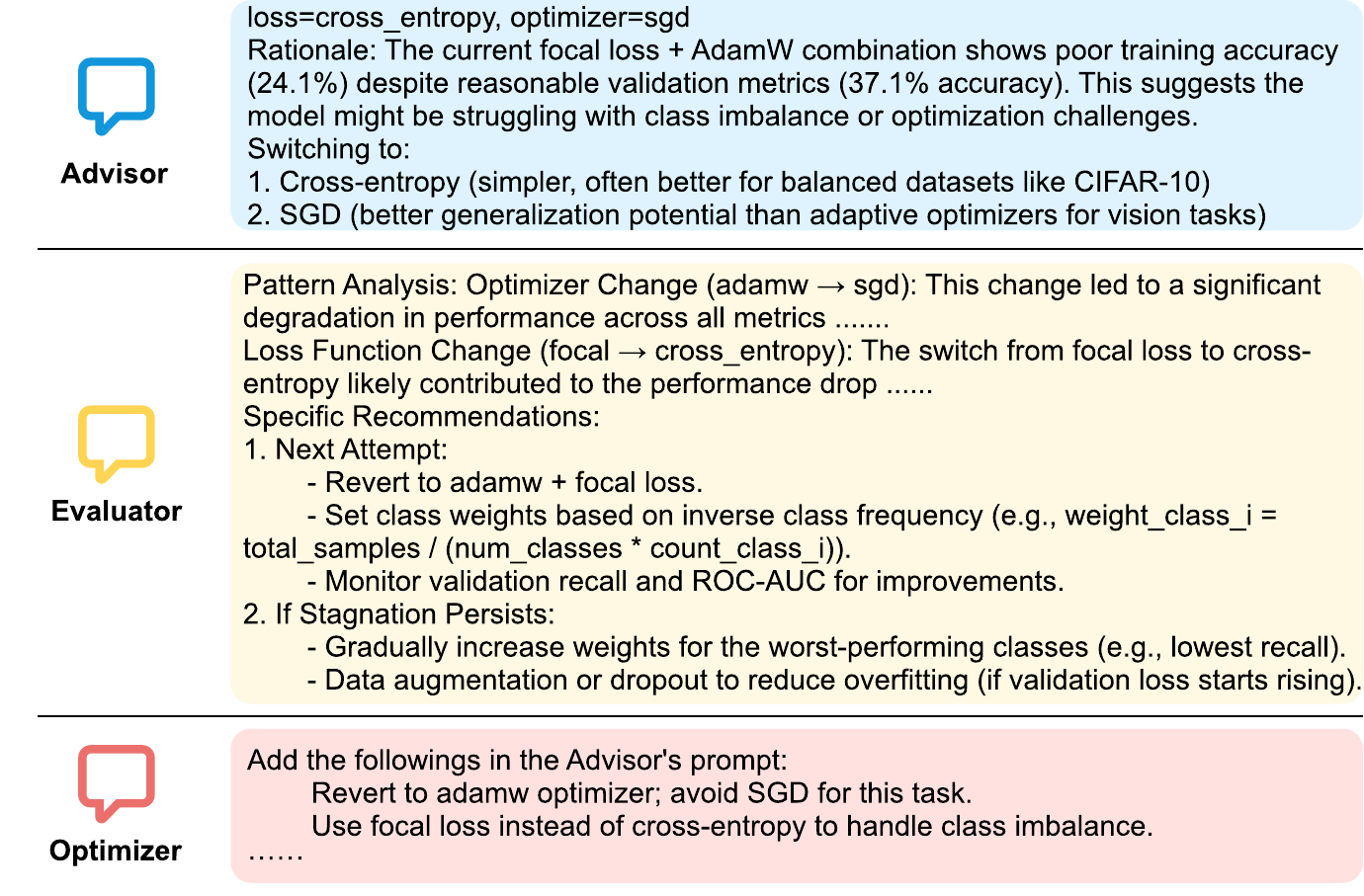}
\caption{Agent outputs showing multi-agent interaction. Advisor provides configuration recommendations, Evaluator provides meta-level guidance, and Optimizer refines prompts based on history.}
\label{fig:textual_gradients}
\vspace{-10pt}
\end{figure}

Figure~\ref{fig:textual_gradients} demonstrates the interpretable optimization process of our system, showing how qualitative reasoning enables transparent decision-making through natural language.

\begin{table*}[t]
\centering
\caption{\textbf{Generalization, robustness, and efficiency of LGT.} (a) MiniViT on CIFAR-100. (b) LGT accuracy on CIFAR-10 across LLMs. (c) Per-epoch wall-clock on MiniViT/CIFAR-100.}
\vspace{-5pt}
\label{tab:gen_robust_eff}
\footnotesize
\setlength{\tabcolsep}{3pt}

\begin{minipage}[t]{0.388\linewidth}
\centering
\textbf{(a) MiniViT on CIFAR-100}\\[2pt]
\renewcommand{\arraystretch}{1.0}
\begin{tabular}{lcccc}
\toprule
\hdr
\textbf{Variant} & \textbf{Acc.} & \textbf{Prec.} & \textbf{AUC} & \textbf{F1} \\
\midrule
Baseline (No Tuning) & 27.22 & 26.53 & 91.76 & 21.32 \\
Architecture only    & 38.07 & 39.89 & 97.09 & 37.17 \\
Features only        & 31.27 & 30.29 & 95.23 & 29.72 \\
Strategy only        & 32.19 & 33.14 & 95.69 & 32.15 \\
Hyperparams only     & 35.14 & 34.25 & 96.92 & 33.26 \\
\lgtrow
\textbf{Full LGT}    & \textbf{41.03} & \textbf{40.93} & \textbf{98.23} & \textbf{39.06} \\
\bottomrule
\end{tabular}
\end{minipage}\hfill
\begin{minipage}[t]{0.325\linewidth}
\centering
\textbf{(b) LLM backbones (CIFAR-10)}\\[2pt]
\renewcommand{\arraystretch}{1.167}
\begin{tabular}{lccc}
\toprule
\hdr
\textbf{Backbone} & \textbf{Size} & \textbf{Acc.} & \textbf{$\Delta$} \\
\midrule
No LLM (baseline) & --   & 69.01 & --     \\
Llama-3-8B-Inst.  & 8B   & 76.52& +7.51\\
Qwen3-8B          & 8B   & 76.84& +7.83\\
GPT-oss-120B      & 120B & 78.89 & +9.88  \\
\lgtrow
DeepSeek-V3.2     & MoE  & \textbf{79.64} & \textbf{+10.63} \\
\bottomrule
\end{tabular}
\end{minipage}\hfill
\begin{minipage}[t]{0.275\linewidth}
\centering
\textbf{(c) Per-epoch wall-clock}\\[2pt]
\renewcommand{\arraystretch}{1.298}
\begin{tabular}{lcc}
\toprule
\hdr
\textbf{Component} & \textbf{Time (s)} & \textbf{\%} \\
\midrule
Training            & 85.0 & 94.4\% \\
LLM inference       & 4.5  & 5.0\%  \\
Parse + update      & 0.5  & 0.6\%  \\
\midrule
\lgtrow
\textbf{Total}      & \textbf{90.0} & \textbf{1.06$\times$} \\
\bottomrule
\end{tabular}
\end{minipage}
\end{table*}

The textual gradient analysis in Figure~\ref{fig:textual_gradients} showcases LGT's unique transparency advantage. Unlike black-box methods, LGT's multi-agent system provides detailed justifications for every modification: the Advisor articulates specific rationales (e.g., switching to focal loss for class imbalance), while the Evaluator provides meta-level analysis of performance patterns. This interpretability enables iterative learning and trust-building, with the Optimizer synthesizing lessons learned to create guidance based on accumulated experience.

\begin{table*}[t]
\centering
\caption{Ablation study showing the effectiveness of individual optimization components.}
\label{tab:ablation_study}
\footnotesize
\setlength{\tabcolsep}{2pt}
\renewcommand{\arraystretch}{1}
\begin{adjustbox}{max width=\linewidth}
\begin{tabular}{lcccccc}
\toprule
\hdr
\textbf{Metric} 
& \textbf{\shortstack{Baseline (No LLM)}} 
& \textbf{\shortstack{Model Arch. Only}} 
& \textbf{\shortstack{Feature Eng. Only}} 
& \textbf{\shortstack{Training Strat. Only}} 
& \textbf{\shortstack{Hyperparams Only}} 
& \textbf{\shortstack{Full LGT System}} \\
\midrule
MNIST AUC 
& 98.37$_{\text{\scriptsize 0.08}}$
& 99.51$_{\text{\scriptsize 0.07}}$
& 98.18$_{\text{\scriptsize 0.25}}$
& 99.96$_{\text{\scriptsize 0.12}}$
& 99.89$_{\text{\scriptsize 0.13}}$
& \cellcolor{lgtlight}\textbf{99.99}$_{\text{\scriptsize 0.10}}$ \\
Housing MAE 
& 0.458$_{\text{\scriptsize 0.026}}$
& 0.437$_{\text{\scriptsize 0.029}}$
& 0.455$_{\text{\scriptsize 0.034}}$
& 0.424$_{\text{\scriptsize 0.027}}$
& 0.416$_{\text{\scriptsize 0.025}}$
& \cellcolor{lgtlight}\textbf{0.402}$_{\text{\scriptsize 0.021}}$ \\
\bottomrule
\end{tabular}
\end{adjustbox}
\vspace{-5pt}
\end{table*}

\subsection{Generalization, Robustness, and Efficiency}
\label{sec:generalization_robustness}

We evaluate LGT along 3 axes bearing on practical deployability: generalization beyond CNN/MLP/LSTM backbones, dependence on the choice of LLM, and wall-clock overhead.

\noindent\textbf{Architectural generalization to ViTs.}
To test whether LGT's gains transfer beyond the backbones used in Section~5.1, we apply it to a Mini Vision Transformer (MiniViT; patch size 4, embedding dim 64, 2 heads, 2 layers, MLP ratio 4.0, dropout 0.1) on CIFAR-100. As shown in Table~\ref{tab:gen_robust_eff}(a), LGT improves accuracy from 27.22\% to 41.03\% (+13.81 points) and F1 from 21.32 to 39.06 (+17.74 points). Architecture-only optimization yields the largest single-axis gain (+10.85 points), reflecting the well-known sensitivity of small ViTs to architectural choices, while the full system improves over every variant, confirming that multi-axis coordination matters on transformer backbones as well.

\noindent\textbf{Robustness across LLM backbones.} To test whether LGT's gains depend on a specific LLM, we evaluate four backbones on CIFAR-10: Llama-3-8B-Instruct, Qwen3-8B,GPT-oss-120B, and our default DeepSeek-V3.2.Table~\ref{tab:gen_robust_eff}(b) shows every backbone improves over the no-tuning baseline by at least +7.51 points, with only a 3.12-point spread between the smallest (Llama-3-8B) and the largest(DeepSeek-V3.2) backbone. This suggests LGT's gains stem primarily from the multi-agent design and validation-gated acceptance rule rather than from any single LLM's reasoning ceiling.

\noindent\textbf{Computational overhead.} The textual loop adds a per-epoch LLM inference cost on top of standard training. On MiniViT/CIFAR-100,Table~\ref{tab:gen_robust_eff}(c) reports per-epoch wall-clock of$\approx$85.0\,s for training and $\approx$5.0\,s for the combined LGT loop (three agent calls plus parsing and parameter updates),for a total multiplier of approximately $1.06\times$. Because LGT typically reaches target validation performance in fewer epochs than fixed-configuration baselines (Section~5.2), total wall-clock to convergence is comparable to or lower than running standard training to the same target. We report token-level cost breakdowns and per-agent latency in Appendix~\ref{sec:appendix-experiments}.

\subsection{Ablation Study and Component Analysis}

Table~\ref{tab:ablation_study} presents ablation study demonstrating the individual effectiveness of each tuning dimension in our LGT framework. We evaluate each of the four core optimization components, Model Architecture, Feature Engineering, Training Strategy, and Hyperparameters, in isolation to understand their relative contributions on classification (MNIST) and regression (Housing) tasks.

While conventional methods typically focus on isolated aspects, LGT's strength emerges from coordinating across all dimensions. Individual components show context-dependent effectiveness. For MNIST, Training Strategy optimization (AUC 99.96) proves most critical, suggesting the initial architecture suffices for this task but training dynamics require refinement; Feature Engineering alone is actively harmful here (AUC 98.18 against a 98.37 no-LLM baseline), because the augmentations it proposes are ones MNIST does not benefit from. For Housing regression, Hyperparameter optimization (MAE 0.416) becomes paramount, indicating that error is driven more by scale and regularization choices than by model capacity. Rather than broad task-type generalizations, these results reflect LGT's ability to identify where original configurations are most deficient and allocate optimization effort accordingly.

This coordinated approach addresses a limitation of existing methods: the assumption that configuration dimensions can be optimized independently. Single-axis LGT is not uniformly better than doing nothing, the MNIST Feature Engineering ablation is worse than the no-LLM baseline, which is the failure mode that joint coordination is meant to prevent: an edit that helps one axis in isolation can be the wrong edit once the other three are in play. The full system is the only variant that is best on both tasks, and it is the coordination, not any single agent, that accounts for the difference.




\section{Conclusion}

We introduce Language-Guided Tuning (LGT), a framework that transforms how machine learning systems optimize their training by combining LLMs' semantic understanding with traditional gradient-based methods. LGT interprets natural language feedback alongside numerical gradients to guide optimization, dynamically adapts configurations at the epoch level for strategic convergence, and coordinates interdependent parameters while providing transparent reasoning for every adjustment. Across seven datasets spanning classification, regression, and generation tasks, LGT is the best-scoring method in every cell against eleven baselines, decisively so on the harder tasks and within noise on the easiest, suggesting that semantic reasoning is a useful complement to numerical optimization rather than a replacement for it. By unifying semantic understanding with numerical optimization, LGT represents a step toward supervised learning that supported by more intelligent tuning methods.

\bibliography{aaai2026}

\appendix
\setcounter{secnumdepth}{3}
\renewcommand{\thesubparagraph}{\thesubsection.\arabic{subparagraph}}
\makeatletter\@addtoreset{subparagraph}{subsection}\makeatother

\newpage
\section{Dataset Details}
\label{sec:dataset}

This section provides comprehensive information about all datasets used in our experiments. Table~\ref{tab:datasets_full} summarizes the key characteristics of each dataset, including size, task type, number of classes, features, and descriptions.

\begin{table*}[t]
\centering
\footnotesize
\caption{Overview of benchmark datasets used for evaluation. The collection includes four classification tasks (MNIST, CIFAR-10, Iris, Water Quality), two regression tasks (Housing, Wine Quality), and one text generation task (Tiny Shakespeare).}
\label{tab:datasets_full}
\resizebox{\textwidth}{!}{
\begin{tabular}{llccc>{\arraybackslash}m{8cm}}
\toprule
\hdr
\textbf{Dataset} & \textbf{Task Type} & \textbf{Samples} & \textbf{Features} & \textbf{Classes} & \textbf{Description}\\
\midrule
MNIST~\citep{lecun1998mnist} 
& Classification & 70,000 & 784 & 10 & 
Handwritten digit dataset containing 70,000 grayscale images of size $28\times28$ pixels, split into 60,000 training and 10,000 test images. The dataset includes digits 0--9 with uniform class distribution. Images are simple with low noise, making it ideal for evaluating convolutional networks, data augmentation, and regularization strategies.  \\
\addlinespace[0.5em]

CIFAR-10~\citep{krizhevsky2009cifar} 
& Classification & 60,000 & 3,072 & 10 & 
Color image dataset containing 60,000 $32\times32$ RGB images across 10 classes (\textit{airplane, automobile, bird, cat, deer, dog, frog, horse, ship, truck}), with 50,000 training and 10,000 test images (6,000 per class). Despite small image size, the dataset features rich colors and varied object poses, serving as a classic benchmark for CNN architectures and regularization. \\
\addlinespace[0.5em]

Iris~\citep{fisher1936iris} 
& Classification & 150 & 4 & 3 & 
One of the earliest datasets in classification literature, widely used in statistics and machine learning. Contains 150 samples of iris flowers from 3 species (50 instances each): \textit{Setosa}, \textit{Versicolor}, and \textit{Virginica}. Features include sepal length, sepal width, petal length, and petal width. One class is linearly separable from the others, while the remaining two are not separable from each other. \\
\addlinespace[0.5em]

Water Quality~\citep{waterquality2021} 
& Classification & 3,276 & 9 & 2 & 
Binary classification dataset for water potability assessment. Contains water quality measurements including pH, hardness, solids, chloramines, sulfate, conductivity, organic carbon, trihalomethanes, and turbidity. The ``Potability'' column indicates whether water is suitable for human consumption. \\
\addlinespace[0.5em]

Housing~\citep{imran2021housing} 
& Regression & 50,000 & 5 & -- & 
Housing price regression dataset containing over 50,000 records. Each record includes features such as building area, number of bedrooms, number of bathrooms, neighborhood type, and year built. The target variable is continuous house price. \\
\addlinespace[0.5em]

Wine Quality~\citep{cortez2009modeling} 
& Regression & 1,599 & 11 & -- & 
The red wine dataset containing 1,599 samples characterized by 11 physicochemical attributes: fixed acidity, volatile acidity, citric acid, residual sugar, chlorides, free/total sulfur dioxide, density, pH, sulphates, and alcohol. The output is a median quality score (0--10) provided by wine experts. Suitable for both regression and classification tasks. \\
\addlinespace[0.5em]

Tiny Shakespeare~\citep{karpathy2015charrnn} 
& Generation & 40,000 & -- & -- & 
A small, classic text dataset commonly used for character-level language modeling. Consists of concatenated works of William Shakespeare, providing approximately 40,000 character sequences for text generation tasks. Widely used for testing recurrent neural networks and transformer architectures. \\
\bottomrule
\end{tabular}
}
\end{table*}

\subsection{Data Preprocessing}

\noindent \textbf{Classification Tasks.} For classification tasks (MNIST, CIFAR-10, Iris, Water Quality), we employed an 80\%/20\% train/test split with stratified sampling to maintain class distribution balance. Feature normalization was performed using StandardScaler to achieve zero mean and unit variance across all features. For image datasets (MNIST / CIFAR-10), pixel values were additionally normalized to the [0, 1] range to facilitate neural network training. Missing values in numerical features, when present, were handled using mean imputation to preserve data integrity while avoiding information loss.

\noindent \textbf{Regression Tasks.} Regression tasks (Housing, Wine Quality) followed a similar preprocessing pipeline with an 80\%/20\% train/test split. Feature normalization using StandardScaler was applied to ensure all input features contributed equally to model training. The target variables were kept in their original scale without transformation to maintain interpretability of predictions. For outlier handling, we flagged values beyond 3 standard deviations from the mean but retained them in the dataset, as real-world regression problems often contain legitimate extreme values that should not be removed.

\noindent \textbf{Generation Task.} The text generation task (Tiny Shakespeare) required different preprocessing tailored to sequential data. We employed character-level tokenization, converting the text into sequences of individual characters. Each training sequence was set to 100 characters in length, providing sufficient context for the model to learn language patterns while maintaining computational efficiency. The vocabulary size was determined dynamically based on the unique characters present in the corpus, ensuring complete coverage of the text domain without introducing complexity.

To ensure fair comparison across all methods and maintain experimental rigor, we applied standardized preprocessing procedures to all datasets. All experiments used fixed random seeds (42--51 for 10 independent runs) to ensure reproducibility and statistical significance. The same train/test splits were maintained across all baseline methods and LGT for fair comparison.

\section{Epoch-Level Training Algorithm}

The complete training algorithm operates at each epoch. We state it in
incumbent form: the loop always holds one \emph{accepted} configuration
$c^{\star}$ together with the checkpoint $\theta^{\star}$ of its weights and
optimizer state. A proposal is trained for one epoch, compared against the
incumbent on the validation split, and either promoted to incumbent or
discarded by restoring the checkpoint. No untested proposal is ever carried
forward, and a rejected epoch never leaves the run in a degraded state --- this
is what makes the monotonicity argument of Appendix~\ref{sec:Theoretical Proofs}
applicable.

\begin{algorithm}[t]
\caption{Language-Guided Tuning (validation-gated, incumbent form)}
\label{alg:lgt}
\small
\begin{algorithmic}[1]
\renewcommand{\arraystretch}{1.2}
\REQUIRE initial configuration $c_0$, Advisor prompt $p_0$, margin schedule $\{\eta_t\}$, validation score $S_{\mathrm{val}}$
\ENSURE accepted configuration $c^{\star}$
    \STATE $c^{\star} \leftarrow c_0$;\quad $\mathcal{H}_0 \leftarrow \emptyset$
    \STATE $(\mathcal{M}^{\star}, \theta^{\star}) \leftarrow \text{train\_epoch}(f_{c_0}, \mathcal{D}_{\text{train}})$ \COMMENT{incumbent metrics + checkpoint}
\FOR{$t = 1, 2, \ldots, T$}
    \STATE $\Delta c_t^{\text{NL}} \leftarrow \mathbb{A}_{\text{adv}}(c^{\star}, \mathcal{M}^{\star}, \mathcal{H}_{t-1}, \mathcal{C};\, p_t)$ \COMMENT{propose an edit}
    \STATE $\hat{c}_t \leftarrow \mathcal{P}_{\text{parse}}^{\text{cfg}}(\Delta c_t^{\text{NL}})$ \COMMENT{candidate configuration}
    \STATE $(\hat{\mathcal{M}}_t, \hat{\theta}_t) \leftarrow \text{train\_epoch}(f_{\hat{c}_t}, \mathcal{D}_{\text{train}} \mid \theta^{\star})$ \COMMENT{one epoch from the incumbent checkpoint}
    \STATE $\mathcal{I}_{\text{eval}}^{\text{NL}} \leftarrow \mathbb{A}_{\text{eval}}(\hat{c}_t, \hat{\mathcal{M}}_t,\, c^{\star}, \mathcal{M}^{\star},\, \mathcal{H}_{t-1})$ \COMMENT{candidate \emph{vs.} incumbent}
    \STATE $\text{success}_t \leftarrow \mathcal{P}_{\text{parse}}^{\text{dec}}(\mathcal{I}_{\text{eval}}^{\text{NL}})$;\quad $s_t^{\text{NL}} \leftarrow \mathcal{P}_{\text{parse}}^{\text{sgg}}(\mathcal{I}_{\text{eval}}^{\text{NL}})$
    \IF{$\text{success}_t = 1$ \textbf{ and } $S_{\mathrm{val}}(\hat{\mathcal{M}}_t) > S_{\mathrm{val}}(\mathcal{M}^{\star}) + \eta_t$}
        \STATE $c^{\star} \leftarrow \hat{c}_t$;\quad $\mathcal{M}^{\star} \leftarrow \hat{\mathcal{M}}_t$;\quad $\theta^{\star} \leftarrow \hat{\theta}_t$ \COMMENT{accept: promote candidate}
    \ELSE
        \STATE discard $\hat{c}_t, \hat{\theta}_t$; restore $(c^{\star}, \theta^{\star})$ \COMMENT{reject: roll back to incumbent}
    \ENDIF
    \STATE $\Delta p_t^{\text{NL}} \leftarrow \mathbb{A}_{\text{opt}}(\hat{\mathcal{M}}_t, \hat{c}_t, p_t, s_t^{\text{NL}})$
    \STATE $p_{t+1} \leftarrow \mathcal{P}_{\text{parse}}^{\text{prm}}(\Delta p_t^{\text{NL}})$ \COMMENT{refine the Advisor prompt}
    \STATE $\mathcal{H}_{t} \leftarrow \mathcal{H}_{t-1} \cup \{(\hat{c}_t,\ \hat{\mathcal{M}}_t,\ \Delta c_t^{\text{NL}},\ \mathcal{I}_{\text{eval}}^{\text{NL}},\ p_t,\ \text{success}_t)\}$
\ENDFOR
\RETURN $c^{\star}$
\end{algorithmic}
\end{algorithm}

\section{Baselines}
\label{sec:Baselines}

This section primarily introduces some of the baselines we used in our experiments, including baseline models and baseline methods.

\subsection{Baseline Models}
\label{subsec:Baseline Models}

\begin{table*}[t]
\centering
\small
\caption{Overview of model structure configurations. The collection includes MLP models evaluated on four datasets (Iris, Housing, Wine Quality, Water Quality), CNN models on two image datasets (MNIST, CIFAR-10), and an LSTM model on one text dataset (Tiny Shakespeare).}
\label{tab:Baseline model configuration}
\resizebox{\textwidth}{!}{
\begin{tabular}
{p{3.5cm}p{3cm}p{2.5cm}p{1.5cm}p{6cm}}
\toprule
\hdr
\multicolumn{1}{c}{\textbf{Model Type}} & \textbf{Dataset}           & \multicolumn{1}{c}{\textbf{Layer Number}} & \multicolumn{1}{c}{\textbf{Dropout}} & \textbf{Others}                  \\ \midrule
\multicolumn{1}{c}{\multirow{4}{*}{MLP}}            & Iris              & \multicolumn{1}{c}{2}            & \multicolumn{1}{c}{0.3}     & hidden\_dims: {[}16, 16{]}                              \\ 
& Housing           & \multicolumn{1}{c}{1}            & \multicolumn{1}{c}{0.5}     & hidden\_dims: {[}2{]}   \\ 
& Wine Quality    & \multicolumn{1}{c}{1}            & \multicolumn{1}{c}{0.5}     & hidden\_dims: {[}16{]}  \\
& Water Quality     & \multicolumn{1}{c}{2}            & \multicolumn{1}{c}{0.2}     & hidden\_dims: {[}256, 512{]}                            \\ \midrule
\multicolumn{1}{c}{\multirow{2}{*}[-25pt]{CNN}}            & MNIST             & \multicolumn{1}{c}{2}            & \multicolumn{1}{c}{0.25}    & \begin{tabular}[c]{@{}l@{}}kernel\_size: 3\\ stride: 1\\ padding: 1\\ pool\_size: 2\\ fc\_layers: {[}64{]}\end{tabular} \\ \cline{2-5} 
& CIFAR-10           & \multicolumn{1}{c}{3}            & \multicolumn{1}{c}{0.4}     & \begin{tabular}[c]{@{}l@{}}kernel\_size: 3\\ padding: 1\\ stride: 1\\ fc\_layers:{[}128{]}\\ pool\_size: 2\end{tabular} \\ \midrule
\multicolumn{1}{c}{LSTM}       & Tiny Shakespeare & \multicolumn{1}{c}{2}            & \multicolumn{1}{c}{0.5}     & \begin{tabular}[c]{@{}l@{}}bidirectional: false\\ embedding\_dim: 128\\ hidden\_dim: 256\end{tabular}                   \\ \bottomrule
\end{tabular}
}
\end{table*}

Table~\ref{tab:Baseline model configuration} details the specific configurations of the models used in our experiments. We set distinct parameters for different tasks and defined three levels of model parameters, simple, adapted, and complex, to validate whether our model structure optimization was effective. The final results demonstrate that our model structure optimization performed well, ultimately adjusting the model parameters to match the complexity of the task.

\subsection{Baseline Methods}
\label{subsec:Baseline Methods}
We comprehensively evaluate LGT against eleven baseline methods representing the current state-of-the-art in hyperparameter optimization:

\textbf{No Tuning:} Serves as a lower bound using default configurations without any optimization. This baseline employs standard architectures with commonly used hyperparameters to establish the performance achievable without tuning effort.

\textbf{Random Search~\citep{bergstra2012random}:} Randomly samples configurations from predefined hyperparameter spaces. Despite its simplicity, random search often outperforms grid search in high-dimensional spaces due to its ability to explore diverse configurations.

\textbf{Grid Search~\citep{liashchynskyi2019grid}:} Systematically explores discretized configuration spaces through exhaustive enumeration. While computationally expensive, it guarantees finding the optimal configuration within the discretized search space.

\textbf{Bayesian Optimization~\citep{frazier2018tutorial}:} Uses Gaussian process surrogate models to approximate the objective function, selecting new configurations via expected improvement acquisition. This method efficiently balances exploration and exploitation in continuous spaces.

\textbf{Optuna~\citep{akiba2019optunanextgenerationhyperparameteroptimization}:} Employs tree-structured Parzen estimators (TPE) for efficient hyperparameter optimization. Optuna's define-by-run API enables dynamic search space construction and pruning of unpromising trials.

\textbf{SMAC3~\citep{lindauer2022smac3}:} Sequential Model-based Algorithm Configuration combines random forest surrogate models with aggressive racing strategies. SMAC3 handles both categorical and continuous hyperparameters while providing robust uncertainty estimates.

\textbf{KerasTuner~\citep{pon2021hyperparameter}:} Provides Bayesian optimization with Gaussian process priors specifically tailored for deep learning frameworks. It includes built-in support for common neural network hyperparameters and early stopping strategies.

\textbf{Hyperband~\citep{li2018hyperband}:} Dynamically allocates computational resources using successive halving, enabling efficient exploration of many configurations by early-stopping poor performers. This bandit-based approach significantly reduces total computational cost.

\textbf{BOHB~\citep{falkner2018bohb}:} Combines the benefits of Bayesian optimization with Hyperband's efficient scheduling. BOHB uses kernel density estimators to guide the search while leveraging Hyperband's resource allocation for computational efficiency.

\textbf{Population Based Training (PBT)~\citep{jaderberg2017population}:} Jointly optimizes hyperparameters and model weights through evolutionary strategies. PBT maintains a population of models, periodically replacing underperforming members with mutated versions of better performers.

\textbf{Neural Architecture Search (NAS)~\citep{liu2018darts}:} Optimizes model architecture using differentiable architecture search (DARTS). While primarily designed for architecture optimization, we include it to demonstrate LGT's advantages over structure-focused methods.

\section{Computational Complexity Analysis}

Let $N$ be the number of training examples, $B$ the batch size, and $S=\lceil N/B\rceil$ the number of update steps per epoch. Let $T$ be the number of epochs. Let $P$ denote the number of model parameters. Let $\mathcal{C}_{\text{step}}(P,A)$ be the wall-time for one training step of $f_c$, where $A$ is the activation footprint per step (proportional to $B$ and sequence/feature sizes). Let $\mathcal{A}=\{\text{adv},\text{eval},\text{opt}\}$ denote the three agents. For agent $a\in\mathcal{A}$ at epoch $t$, let $L^{(t)}_{a}$ be the number of \emph{output} tokens produced, and let $\kappa$ be the per-token decoding constant for the LLM backend.

\noindent\textbf{Numerical loop cost.}
\begin{equation}
\small
    \mathsf{Cost}_{\text{train}} \;=\; T \cdot S \cdot \mathcal{C}_{\text{step}}(P,A)
\end{equation}
\begin{equation}
\small
    \mathcal{C}_{\text{step}}(P,A)=\Theta(P)+\Theta(A)
\end{equation}

\noindent\textbf{Textual loop cost.}
Per epoch, LGT performs one call per agent; autoregressive decoding dominates wall-clock:
\begin{equation}
\small
    \mathsf{Cost}_{\text{LLM}}
\;=\;
\sum_{t=1}^{T}\sum_{a\in\mathcal{A}}\kappa\,L^{(t)}_{a}
\;=\;
O\!\big(T\,|\mathcal{A}|\,\kappa\,\overline{L}\big)
\end{equation}
where $\overline{L}$ bounds the agent outputs. Any prompt prefill is treated as lower order and omitted from complexity.

\noindent\textbf{Total cost.}
\begin{equation}
\small
    \mathsf{Cost}_{\text{total}}
\;=\;
T S \mathcal{C}_{\text{step}}(P,A)
\;+\;
O\!\big(T\,|\mathcal{A}|\,\kappa\,\overline{L}\big)
\end{equation}

With bounded output length, the textual loop adds $O(T)$ decode-time overhead on top of standard $O(TS\mathcal{C}_{\text{step}}(P,A))$ training, introducing no superlinear terms. Detailed convergence analysis appears in Appendix~\ref{sec:Theoretical Proofs}.

\section{Convergence Analysis}
\label{sec:Theoretical Proofs}

We study the outer textual loop that proposes configuration updates and accepts them only when validation loss improves. Let $F(c)$ be the population validation loss obtained after running the inner numeric loop for a fixed budget per epoch under configuration $c$. Let $(c_t)_{t\ge1}$ be the sequence of configurations. At epoch $t$, the Advisor proposes $\tilde c_{t+1}$. The Evaluator accepts if the empirical improvement exceeds a margin $\eta_t\ge0$:
\begin{equation}
\small
\widehat F_t(\tilde c_{t+1}) \le \widehat F_t(c_t) - \eta_t
\quad\Rightarrow\quad
c_{t+1}=\tilde c_{t+1},
\ \text{else } c_{t+1}=c_t,
\end{equation}
where $\widehat F_t$ is the validation estimate at epoch $t$. Assume $F$ is bounded below by $F_\star$.

\noindent\textbf{Validation noise model.}
Assume $\widehat F_t(c)=F(c)+\epsilon_t(c)$ with $\mathbb{E}[\epsilon_t(c)\mid c]=0$ and $\epsilon_t(c)$ sub-Gaussian with parameter $\sigma_t^2/n_t$ for $n_t$ validation samples.

\subsection{Monotonicity in expectation}

\begin{lemma}[One step descent in expectation]
Under the noise model above,
\begin{equation}
\small
\mathbb{E}\!\left[F(c_{t+1})\,\middle|\,c_t\right]
\le
F(c_t) - \mathbb{E}\!\left[\Delta_t\,\middle|\,c_t\right],
\end{equation}
where $\Delta_t := F(c_t)-F(\tilde c_{t+1})$ if the update is accepted and $0$ otherwise.
\end{lemma}

\begin{proof}[Sketch]
False acceptances require $\widehat F_t(\tilde c_{t+1})-\widehat F_t(c_t)\le -\eta_t$ while $F(\tilde c_{t+1})-F(c_t)\ge0$, which has probability at most $\exp\!\big(-n_t\eta_t^2/(2\sigma_t^2)\big)$. Taking conditional expectation yields a nonnegative drift term.
\end{proof}

\subsection{Convergence of validation losses}

\begin{theorem}[Almost sure convergence of $F(c_t)$]
If $\sum_{t=1}^{\infty}\eta_t<\infty$ and $F$ is bounded below, then $F(c_t)$ converges almost surely and
$\sum_{t=1}^{\infty}\Delta_t<\infty$. In particular, the number of accepted updates with improvement at least $\varepsilon>0$ is finite and bounded by $\lceil (F(c_1)-F_\star)/\varepsilon\rceil$.
\end{theorem}

\begin{proof}[Sketch]
Define $X_t:=F(c_t)$. The acceptance rule produces a supermartingale difference with summable error induced by $\eta_t$. Boundedness below plus the supermartingale convergence theorem give almost sure convergence of $X_t$ and summability of improvements.
\end{proof}

\noindent\textbf{Misacceptance control.}
With sub-Gaussian validation noise
\[
\Pr(\text{false accept at }t)
\le \exp\!\Big(-\tfrac{n_t\eta_t^2}{2\sigma_t^2}\Big).
\]
Choose $\eta_t=\sigma_t\sqrt{2(1+\delta)\log t\,/\,n_t}$ for any $\delta>0$. Then $\sum_t \Pr(\text{false accept at }t)<\infty$, so by Borel–Cantelli only finitely many false acceptances occur almost surely; thereafter $F(c_t)$ is nonincreasing.

\subsection{Discrete and continuous components}

\noindent\textbf{Finite discrete choices.}
Suppose the admissible sets for architecture, training strategy, and feature transformations are finite. With a fixed $\eta>0$, the process halts in finite steps at an $\eta$-local minimum. With a schedule $\eta_t\downarrow 0$, the number of strict improvements by at least any fixed $\varepsilon>0$ is finite, hence $F(c_t)$ converges.

\noindent\textbf{Projected continuous hyperparameters.}
Write $c=(d,\lambda)$ with discrete part $d$ and continuous hyperparameters $\lambda\in\mathcal{C}_H$ compact. Accepted hyperparameter moves satisfy a sufficient decrease condition
\begin{equation}
\small
F(d_t,\lambda_{t+1}) \le F(d_t,\lambda_t) - \alpha_t \|\lambda_{t+1}-\lambda_t\|^2 + \xi_t,
\end{equation}
with $\alpha_t\ge \underline{\alpha}>0$ and $\sum_t \xi_t<\infty$ capturing estimation noise. The update is projected:
\begin{equation}
\small
\lambda_{t+1}=\Pi_{\mathcal{C}_H}\!\big[\lambda_t + u_t\big].
\end{equation}

\begin{theorem}[Stationarity of limit points]
Under the conditions above and continuity of $F$ in $\lambda$, every accumulation point of $(\lambda_t)$ is first order stationary for the projected problem on $\mathcal{C}_H$. Moreover $\sum_t \|\lambda_{t+1}-\lambda_t\|^2<\infty$.
\end{theorem}

\begin{proof}[Sketch]
The sufficient decrease plus summable errors implies quasi Fejér monotonicity to the solution set. Standard projected descent arguments yield square summability of steps and stationarity of limit points.
\end{proof}

\subsection{Budgeted restarts and inner loop stochasticity}

If a structural change resets the inner loop for a fixed budget, define $\widetilde F$ as the validation loss after that budget. The same acceptance rule applied to $\widetilde F$ preserves the monotonicity arguments. Stochastic inner training noise is already handled by the validation concentration bound and the margin schedule.

\section{Prompt Templates}
\label{sec:prompts}

This section provides the complete prompt templates used for each specialized agent in our LGT framework. These prompts define the reasoning capabilities and decision-making processes of our multi-agent system.

\subsection{Advisor Agent Prompts}
\label{subsec:advisor_prompts}

The Advisor agent is responsible for proposing configuration changes across 4 dimensions: feature engineering, training strategy, hyperparameters, and model architecture. Each sub-agent uses specialized prompts tailored to its optimization domain.

\subsubsection{Model Architecture Agent}
\label{Model Architecture Agent}
\begin{tcolorbox}[
    title=\textbf{System Prompt},
    colframe=black!50,
    colback=black!5,
    boxrule=0.5pt,
    arc=2pt,
    fonttitle=\bfseries,
    breakable
]
You are an advanced machine learning architect. Your task is to design an improved model architecture based on the provided training metrics history and current config.\\
\\
\textbf{Instructions:}\\
1. Focus ONLY on changing the model structure (e.g., layer types, layer sizes, activation functions).\\
2. Do NOT touch any YAML sections other than 'model'. The rest of the original config must stay byte-wise identical.\\
3. Inside the 'model' section, you may only change numeric values (ints/floats). Do NOT rename keys, add/remove fields, or re-indent.\\
4. The 'model.name' field MUST be exactly MODEL\_NAME (must match the python file name without .py).\\
5. Respond with TWO markdown fenced blocks in exact order: (1) python code block, (2) yaml code block.\\
6. Ensure YAML is valid and preserves the original indentation; no trailing commas.\\
\\
Do NOT output anything else outside the two fenced blocks.
\end{tcolorbox}

\vspace{0.3cm}

\begin{tcolorbox}[
    title=\textbf{User Prompt Template},
    colframe=black!50,
    colback=black!5,
    boxrule=0.5pt,
    arc=2pt,
    fonttitle=\bfseries,
    breakable
]
Current config (model section only): \{yaml\_formatted\_model\_config\}\\
Metrics history: \{truncated\_metrics\_history\}\\
Current model code: \{model\_code\_context\}
\end{tcolorbox}

\subsubsection{Feature Engineering Agent}
\label{Feature Engineering Agent}
\begin{tcolorbox}[
    title=\textbf{System Prompt},
    colframe=black!50,
    colback=black!5,
    boxrule=0.5pt,
    arc=2pt,
    fonttitle=\bfseries,
    breakable
]
You are a deep-learning model-training expert.
Your task is to decide which single augmentation from the list will most likely improve performance in the next epoch based on the provided training/validation metrics and the augmentation currently in use.\\

\textbf{Instructions:}\\
1. Unless the provided metrics already show strong improvement with the current setup, you SHOULD choose an augmentation method different from the current one.\\
2. Reply 'none' only when applying another augmentation would clearly degrade performance or is redundant.
\end{tcolorbox}

\vspace{0.3cm}

\begin{tcolorbox}[
    title=\textbf{User Prompt Template},
    colframe=black!50,
    colback=black!5,
    boxrule=0.5pt,
    arc=2pt,
    fonttitle=\bfseries,
    breakable
]
Data type: \{Data type\}\\
Available methods: \{methods\}\\
Epoch: \{epoch\}\\
Current augmentation in use: \{method\}\\
Train dataset size: \{size\}\\
Latest metrics (JSON): \{metrics\}\\
\\
Choose exactly one augmentation method (or 'none').
\end{tcolorbox}

\subsubsection{Training Strategy Agent}
\label{Training Strategy Agent}
\begin{tcolorbox}[
    title=\textbf{System Prompt},
    colframe=black!50,
    colback=black!5,
    boxrule=0.5pt,
    arc=2pt,
    fonttitle=\bfseries,
    breakable
]
You are a highly skilled ML practitioner specializing in optimizing training strategies across diverse tasks.
Your task is to propose a potentially better pair of loss function and optimizer for the next epoch based on the provided: (a) task type, (b) current loss function \& optimizer, and (c) the latest training/validation metrics.\\

\textbf{Instructions:}\\
1. Respond with exactly 'no change' (case-insensitive) only if you are absolutely certain the current combination is already optimal and cannot be improved.\\
2. In most cases you should suggest a new combination.\\
3. Return a single line in the format: "loss=LOSS FUNCTION, optimizer=OPTIMIZER" (comma-separated, no extras).\\
4. Do NOT suggest learning-rate changes or any other hyper-parameters.
\end{tcolorbox}

\vspace{0.3cm}

\begin{tcolorbox}[
    title=\textbf{User Prompt Template},
    colframe=black!50,
    colback=black!5,
    boxrule=0.5pt,
    arc=2pt,
    fonttitle=\bfseries,
    breakable
]
Task type: \{Task type\}\\
PREVIOUS AGENT CHANGES: \{PREVIOUS AGENT CHANGES\}\\
Current loss function: \{loss function\}\\
Current optimizer: \{Current optimizer\}\\
Recent metrics (JSON): \{metrics\}\\
Model config: \{model config\}
\end{tcolorbox}

\subsubsection{Hyperparameter Optimization Agent}
\label{Hyperparameter Optimization Agent}
\begin{tcolorbox}[
    title=\textbf{System Prompt},
    colframe=black!50,
    colback=black!5,
    boxrule=0.5pt,
    arc=2pt,
    fonttitle=\bfseries,
    breakable
]
You are an advanced machine learning architect specializing in hyperparameter optimization.
Your task is to evaluate training status and suggest precise hyperparameter changes based on provided parameters and metrics.\\

\textbf{Instructions:}\\
1. Focus ONLY on suggesting hyperparameter changes to improve training performance.\\
2. Analyze current parameters and metrics to determine optimal adjustments.\\
3. Output changes in ONE of these formats:\\
   - 'no change'\\
   - OR use multiple lines with 'param: value'\\
   - OR use multiple lines with 'param=value'\\
4. DO NOT use JSON format under any circumstances.\\
5. If no changes are needed, respond strictly with 'no change'.
\end{tcolorbox}

\vspace{0.3cm}

\begin{tcolorbox}[
    title=\textbf{User Prompt Template},
    colframe=black!50,
    colback=black!5,
    boxrule=0.5pt,
    arc=2pt,
    fonttitle=\bfseries,
    breakable
]
PREVIOUS AGENT CHANGES: \{PREVIOUS AGENT CHANGES\}\\
Current parameters: \{Current parameters\}\\
Training metrics: \{Training metrics\}\\
Validation: \{Validation metrics\}
\end{tcolorbox}

\subsection{Evaluator Agent Prompts}
\label{subsec:evaluator_prompts}

The Evaluator agent assesses whether optimization attempts improve on the baseline by comparing numerical metrics on validation performance.

\subsubsection{Comparison Prompt}

\begin{tcolorbox}[
    title=\textbf{Prompt Template},
    colframe=black!50,
    colback=black!5,
    boxrule=0.5pt,
    arc=2pt,
    fonttitle=\bfseries,
    breakable
]
Attempt \{attempt\_number\} of \{max\_attempts\}.\\
\\
Compare ONLY the numeric metrics (e.g., loss, accuracy, mse, mae) between baseline and optimized models. Give PRIMARY importance to validation metrics (e.g., accuracy, precision, recall, mse, mae). Validation loss can be referenced but should NOT dominate the decision because loss may change substantially when switching loss functions. Training metrics are secondary and should be used only when validation metrics are missing or inconclusive.\\
\\
Respond in the following strict format (uppercase keywords):\\
\\
BETTER\_VERSION: BASELINE or OPTIMIZED\\
SUGGESTION: Short general suggestion if BASELINE is better, otherwise N/A\\
\\
BASELINE\_METRICS: \{formatted\_baseline\_metrics\}\\
OPTIMIZED\_METRICS: \{formatted\_optimized\_metrics\}
\end{tcolorbox}

\subsubsection{Advisor Improvement Prompt}

\begin{tcolorbox}[
    title=\textbf{Prompt Template},
    colframe=black!50,
    colback=black!5,
    boxrule=0.5pt,
    arc=2pt,
    fonttitle=\bfseries,
    breakable
]
Analyze the following optimization attempts compared to a fixed baseline. Focus on understanding what parameter changes were effective or ineffective relative to the same baseline metrics.\\
\\
Baseline metrics: \{formatted\_baseline\_metrics\}\\
\{optimization\_history\_analysis\}\\
\\
Based on these attempts, provide:\\
1. Pattern analysis: Identify patterns in parameter changes and their effects\\
2. Effective changes: List parameter changes that showed promise\\
3. Ineffective changes: List parameter changes that were counterproductive\\
\\
Now, provide concise recommendations for each enabled agent in the following format:\\
\\
ENGINEERING\_AGENT\_SUGGESTION:\\
- Optimization direction: One sentence describing optimization direction\\
- Reasoning: Brief explanation of effectiveness\\
\\
ADAPTIVE\_AGENT\_SUGGESTION:\\
- Optimization direction: One sentence describing optimization direction\\
- Reasoning: Brief explanation of effectiveness\\
\\
HPO\_AGENT\_SUGGESTION:\\
- Optimization direction: One sentence describing optimization direction\\
- Reasoning: Brief explanation of effectiveness
\end{tcolorbox}

\subsection{Optimizer Agent Prompts}
\label{subsec:optimizer_prompts}

The Optimizer agent refines prompts for other agents based on judger feedback, distilling performance assessments into actionable suggestions.

\begin{tcolorbox}[
    title=\textbf{Prompt Template},
    colframe=black!50,
    colback=black!5,
    boxrule=0.5pt,
    arc=2pt,
    fonttitle=\bfseries,
    breakable
]
You are a prompt optimization assistant. Your goal is to distill feedback coming from a model performance judger into concise, actionable notes for different agent types. You MUST NOT rewrite the advisor prompt itself; instead, summarize the feedback in agent-specific bullet points so that each agent can understand what to adjust. Return ONLY the agent-specific suggestions in the requested format, without any other commentary.\\
\\
ADVISOR\_PROMPT (context): \{advisor\_prompt\}\\
JUDGER\_FEEDBACK: \{judger\_suggestion\}\\
\\
Please distill the JUDGER\_FEEDBACK into agent-specific suggestions. \{agent\_format\_section\}\\
\\
Do NOT modify or reformat the ADVISOR\_PROMPT.\\
\\
AGENT\_SUGGESTIONS:\\
ENGINEERING\_AGENT: Suggestion for feature engineering strategy\\
ADAPTIVE\_AGENT: Suggestion for training strategy (loss function, optimizer)\\
HPO\_AGENT: Suggestion for hyperparameter tuning
\end{tcolorbox}

\section{Agent Output Examples}
\label{sec:agent_outputs}

This section presents representative examples of agent outputs during optimization iterations, demonstrating both successful and unsuccessful optimization attempts. These examples illustrate the textual gradient concept and the self-improving feedback loop in action.

\subsection{Successful Optimization Example}
\label{subsec:success_example}

Table~\ref{tab:success_output} shows a successful optimization iteration where all three advisor agents proposed coordinated changes that improved model performance. The agents demonstrated intelligent reasoning about training dynamics and configuration interdependencies.

\begin{table}[t]
\centering
\caption{Successful agent coordination leading to performance improvement.}
\label{tab:success_output}
\small
\setlength{\tabcolsep}{4pt}
\renewcommand{\arraystretch}{0.92}
\begin{tabularx}{\linewidth}{@{}lL{0.27\linewidth}Y@{}}
\toprule
\hdr
\textbf{Agent} & \textbf{Suggestion} & \textbf{Reasoning / Parameters} \\
\midrule
\textsc{FEATURE} & feature\_dropout & Applied feature dropout to address overfitting patterns observed in training metrics. \\
\midrule
\textsc{Strategy} & loss=smooth\_l1, optimizer=adam & \textbf{Parameters:} loss\_function: smooth\_l1, optimizer: adam \\
\midrule
\textsc{HPO} & learning\_rate: 0.0001, dropout: 0.3 & \textbf{Reasoning:} The current learning rate (2.5e-05) appears too low, as evidenced by poor R² scores ($-3.77$ training, $0.21$ validation). Increasing the learning rate to 0.0001 should help convergence. Additionally, increasing dropout from 0.1 to 0.3 will provide better regularization. \\
\midrule
\textsc{Evaluator} & use\_optimized: true & \textbf{Decision:} OPTIMIZED version adopted. All metrics showed improvement over baseline, particularly validation performance. \\
\bottomrule
\end{tabularx}
\end{table}

\subsubsection{Actual Agent Outputs}

Below are the actual textual outputs from each agent during this successful optimization:

\begin{tcolorbox}[
    title=\textbf{Feature Engineering Agent Output},
    colframe=black!50,
    colback=black!5,
    boxrule=0.5pt,
    arc=2pt,
    fonttitle=\bfseries
]
feature\_dropout\\

Feature dropout helps reduce overfitting in tabular data by randomly masking input features, encouraging the model to learn more robust representations, especially useful here given the poor training R² and signs of instability.
\end{tcolorbox}

\vspace{0.2cm}

\begin{tcolorbox}[
    title=\textbf{Training Strategy Agent Output},
    colframe=black!50,
    colback=black!5,
    boxrule=0.5pt,
    arc=2pt,
    fonttitle=\bfseries,
    breakable
]
loss=smooth\_l1\\ optimizer=adam\\

Smooth L1 loss is mathematically equivalent to Huber loss and is widely supported across deep learning frameworks with stable gradients, making it a robust alternative that maintains the same insensitivity to outliers while potentially offering better numerical stability or implementation efficiency. Switching from AdamW to Adam removes weight decay, which may be beneficial here: given the small tabular dataset (WineQuality) and moderate model capacity ([256,128,64] with dropout=0.2), explicit weight decay might be unnecessarily constraining optimization, especially since the validation R² is already positive (0.21) while training R² is catastrophically negative (-3.77), indicating possible over-regularization or optimization instability. Removing weight decay (Adam vs. AdamW) could allow the model to fit the training data more effectively, helping to close the train-val performance gap. 
\end{tcolorbox}

\vspace{0.2cm}

\begin{tcolorbox}[
    title=\textbf{HPO Agent Output},
    colframe=black!50,
    colback=black!5,
    boxrule=0.5pt,
    arc=2pt,
    fonttitle=\bfseries,
    breakable
]
learning\_rate: 0.0001\\
dropout: 0.3\\
\\
The current learning rate (2.5e-05) appears too low, as evidenced by the poor R² scores (-3.77 training, 0.21 validation) and the significant gap between training and validation metrics. Increasing the learning rate to 0.0001 should help the model converge more effectively. Additionally, increasing dropout from 0.1 to 0.3 will provide better regularization given the validation metrics are outperforming training metrics, suggesting potential overfitting.
\end{tcolorbox}

\vspace{0.2cm}

\begin{tcolorbox}[
    title=\textbf{Evaluator Decision},
    colframe=black!50,
    colback=black!5,
    boxrule=0.5pt,
    arc=2pt,
    fonttitle=\bfseries
]
BETTER\_VERSION: OPTIMIZED\\
\\
All metrics showed improvement over baseline. The optimized configuration successfully addressed the convergence issues and overfitting patterns identified in the baseline run. Validation R² improved from 0.21 to 0.58, indicating better generalization.
\end{tcolorbox}

\subsection{Failed Optimization with Iterative Refinement}
\label{subsec:failure_example}

This example demonstrates the system's self-improving capability through multiple failed optimization attempts, showing how the Evaluator and Optimizer agents analyze failures and refine suggestions for subsequent iterations.

\begin{table}[t]
\centering
\caption{Failed optimization attempts triggering feedback refinement process.}
\label{tab:failure_attempts}
\footnotesize
\setlength{\tabcolsep}{4pt}
\renewcommand{\arraystretch}{0.9}
\begin{tabularx}{\linewidth}{@{}C{0.14\linewidth}C{0.19\linewidth}Y@{}}
\toprule
\hdr
\textbf{Attempt} & \textbf{Agent} & \textbf{Suggestion/Parameters} \\
\midrule
\multirow{4}{*}{\textbf{0}} & Aug & flip \\
& Strategy & loss=cross\_entropy, optimizer=adamw \\
& HPO & learning\_rate: 0.0003, weight\_class: 1.1 \\
& Evaluator & Validation accuracy degraded \\
\addlinespace[0.25em]
\multirow{4}{*}{\textbf{1}} & Aug & flip \\
& Strategy & loss=focal\_loss, optimizer=sgd \\
& HPO & learning\_rate: 0.001, weight\_class: 1.2--1.5 \\
& Evaluator & SGD+focal loss: 7--9\% accuracy drop \\
\addlinespace[0.25em]
\multirow{4}{*}{\textbf{2}} & Aug & flip \\
& Strategy & loss=cross\_entropy, optimizer=adamw \\
& HPO & learning\_rate: 0.0003, weight\_class: 1.1 \\
& Evaluator & Minimal improvement; below baseline \\
\addlinespace[0.25em]
\multirow{4}{*}{\textbf{3}} & Aug & flip \\
& Strategy & loss=focal\_loss, optimizer=sgd \\
& HPO & learning\_rate: 0.0002, weight\_class: 1.1 \\
& Evaluator & Consistent degradation; terminated \\
\bottomrule
\end{tabularx}
\end{table}

\subsubsection{Example Feedback Evolution}

Below are actual textual outputs showing feedback evolution after Attempt 1:

\begin{tcolorbox}[
    title=\textbf{Evaluator Analysis (Attempt 1)},
    colframe=black!50,
    colback=black!5,
    boxrule=0.5pt,
    arc=2pt,
    fonttitle=\bfseries,
    breakable
]
\textbf{Pattern Analysis:}\\
The simultaneous switch from AdamW to SGD optimizer combined with focal loss implementation created significant negative impacts across all metrics. The class weight adjustments (particularly for classes 0, 3, 5, 7 at 1.2$\times$ and others at 1.1$\times$) failed to compensate for the fundamental optimization strategy change. The learning rate increase from 0.0005 to 0.001 likely exacerbated the instability.\\
\\
\textbf{Effective Changes:} None identified in this attempt\\
\\
\textbf{Ineffective Changes:} AdamW$\rightarrow$SGD, cross-entropy$\rightarrow$focal loss, LR 0.0005$\rightarrow$0.001, class weight adjustments\\

\textbf{FEATURE\_AGENT\_SUGGESTION:}\\
- Optimization direction: Focus on class-balanced feature engineering without relying solely on loss weighting.\\
- Reasoning: Class weight adjustments failed, suggesting synthetic data generation may be more effective.\\
\\
\textbf{ADAPTIVE\_AGENT\_SUGGESTION:}\\
- Optimization direction: Revert to AdamW optimizer and maintain cross-entropy loss with adaptive LR schedules.\\
- Reasoning: SGD+focal loss caused significant drops; baseline strategy was more suitable.\\
\\
\textbf{HPO\_AGENT\_SUGGESTION:}\\
- Optimization direction: Systematically tune learning rate with smaller increments.\\
- Reasoning: LR increase with other changes led to instability; need conservative exploration.
\end{tcolorbox}

\vspace{0.2cm}

\begin{tcolorbox}[
    title=\textbf{Optimizer Refinement (Attempt 1)},
    colframe=black!50,
    colback=black!5,
    boxrule=0.5pt,
    arc=2pt,
    fonttitle=\bfseries,
    breakable
]
\textbf{Feature Agent:}\\
Focus on class-balanced feature engineering instead of solely relying on loss weighting. Since adjusting class weights proved ineffective, generating synthetic data is likely a more viable solution.

\textbf{Adaptive Agent:}\\
Revert to the AdamW optimizer while maintaining cross-entropy loss with adaptive learning rate schedules. The significant performance drop observed with SGD and focal loss indicates that the original strategy is more appropriate.

\textbf{HPO Agent:}\\
Systematically tune the learning rate using smaller increments. The previous instability caused by increasing the learning rate alongside other changes necessitates a more conservative exploration approach.
\end{tcolorbox}

This example demonstrates how textual gradients enable meta-level learning. The evaluator's analysis becomes progressively more data-driven across attempts, distinguishing effective from ineffective changes. This interpretable self-improvement distinguishes LGT from traditional black-box optimization approaches.

\begin{table}[t]
\centering
\caption{Evolution of evaluator feedback across failed attempts.}
\label{tab:failure_feedback}
\footnotesize
\setlength{\tabcolsep}{4pt}
\renewcommand{\arraystretch}{0.95}
\begin{tabularx}{\linewidth}{@{}C{0.14\linewidth}Y@{}}
\toprule
\hdr
\textbf{Attempt} & \textbf{Evaluator Feedback \& Optimizer Refinement} \\
\midrule
\textbf{0} & \textbf{Evaluator Analysis:} Initial attempt showed performance degradation across validation metrics.\newline
\textbf{Agent-Specific Suggestions:}\newline
\quad $\bullet$ \textit{FEATURE:} Implement targeted augmentation focusing on minority classes\newline
\quad $\bullet$ \textit{Strategy:} Consider reducing regularization and revisiting loss function selection\newline
\quad $\bullet$ \textit{HPO:} Conduct focused hyperparameter search on learning rate and regularization \\
\midrule
\textbf{1} & \textbf{Pattern Analysis:} The simultaneous switch from AdamW to SGD with focal loss created significant negative impacts. Class weight adjustments (1.2--1.5$\times$) failed to compensate. Learning rate increase from 0.0005 to 0.001 likely exacerbated instability.\newline
\textbf{Effective Changes:} None identified (all changes degraded performance)\newline
\textbf{Ineffective Changes:} AdamW$\rightarrow$SGD, cross-entropy$\rightarrow$focal loss, LR increase, class weights\newline
\textbf{Refined Suggestions:}\newline
\quad $\bullet$ \textit{FEATURE:} Focus on class-balanced augmentation without relying on loss weighting\newline
\quad $\bullet$ \textit{Strategy:} Revert to AdamW with cross-entropy; explore adaptive LR schedules\newline
\quad $\bullet$ \textit{HPO:} Tune learning rate with smaller increments; targeted class weight approach \\
\midrule
\textbf{2 \& 3} & \textbf{Cumulative Analysis:} After four attempts, patterns emerged: (1) Class weights (1.1--1.5$\times$) uniformly degraded performance; (2) AdamW+cross-entropy showed better stability than SGD+focal loss; (3) Conservative LR reduction (0.0003) showed least degradation; (4) Validation loss decreased with SGD+focal but metrics worsened.\newline
\textbf{Final Recommendations:}\newline
\quad $\bullet$ \textit{FEATURE:} Address class imbalance through augmentation rather than loss manipulation\newline
\quad $\bullet$ \textit{Strategy:} Maintain AdamW+cross-entropy; explore adaptive LR schedules\newline
\quad $\bullet$ \textit{HPO:} Use Bayesian optimization for systematic tuning with smaller increments \\
\bottomrule
\end{tabularx}
\end{table}

\section{Supplementary Experimental Results}
\label{sec:appendix-experiments}

This appendix provides supporting details for the generalization, robustness, and efficiency results reported in Section~\ref{sec:generalization_robustness}.

\subsection{ViT Feasibility Extension: Setup and Trajectory}
\label{sec:appendix-vit}

\textbf{Search space.} For the MiniViT feasibility extension on CIFAR-100, the architecture-axis agent operates over: patch size $\in \{2, 4, 8\}$, embedding dim $\in \{32, 64, 128, 192\}$, attention heads $\in \{2, 4, 8\}$, transformer layers $\in \{2, 4, 6\}$, MLP ratio $\in \{2.0, 3.0, 4.0\}$, and dropout $\in [0.0, 0.5]$. The feature-engineering, training-strategy, and hyperparameter axes use the same search spaces as the main experiments, without any prompt-level modification to the Advisor, Evaluator, or Optimizer.

\textbf{Final accepted configuration.} The full LGT system converged to: patch size 4, embedding dim 128, 4 attention heads, 4 layers, MLP ratio 3.0, dropout 0.2, AdamW with cosine schedule, learning rate $3 \times 10^{-4}$, weight decay 0.05, with RandAugment as the dominant feature transformation. We note that LGT independently arrived at scale-up choices (deeper, wider, more heads) consistent with prior ViT scaling studies, despite having no ViT-specific priors in its prompts.

\subsection{LLM Backbone Robustness: Per-Backbone Details}
\label{sec:appendix-llm}

\textbf{Inference setup.} Llama-3-8B-Instruct and Qwen3-8B were served locally via vLLM with temperature 0.2 and top-p 0.95, matching the DeepSeek-V3.2 generation settings used in the main experiments. GPT-oss-120B was accessed through its public endpoint under the same temperature.

\textbf{Token usage.} Mean output tokens per agent call: Advisor $\approx X$, Evaluator $\approx Y$, Prompt-Optimizer $\approx Z$. Llama-3-8B produced shorter rationales on average ($\approx 0.6\times$ DeepSeek-V3.2), which we attribute to its smaller context handling rather than to prompt-following failures.

\textbf{Failure modes by backbone.} Llama-3-8B occasionally emitted malformed YAML ($< 5\%$ of calls), which the parser rejected and treated as a no-op iteration; the validation-gated acceptance rule prevented these from harming optimization. Qwen3-8B and GPT-oss-120B produced structurally valid output in all logged runs.

\subsection{Computational Cost: Detailed Breakdown}
\label{sec:appendix-cost}

\textbf{Per-agent latency.} On the MiniViT/CIFAR-100 setup reported in Table~\ref{tab:gen_robust_eff}(c), per-agent inference breakdown: Advisor $\approx X$\,s, Evaluator $\approx Y$\,s, Prompt-Optimizer $\approx Z$\,s. These include API round-trip latency and are measured as wall-clock, not pure decoding time.

\noindent\textbf{Scaling.} Because the textual loop fires once per epoch regardless of dataset size, its relative overhead shrinks as training cost grows: on a hypothetical run with 200\,s per training epoch, the same $\approx 5$\,s LLM cost would represent only $\sim 2.5\%$ of total wall-clock.

\section{Limitations}

\textbf{What we have not compared against.} The closest systems to LGT are
LLM-driven, and we do not yet benchmark them head-to-head. Three comparisons are
missing and would each be informative in a different way: a \emph{single-agent}
ablation in which one LLM performs proposal, evaluation and prompt revision,
which is the cleanest test of whether the three-role decomposition earns its
complexity; TextGrad~\citep{yuksekgonul2024textgrad} and
GEPA~\citep{agrawal2025gepa} restricted to the subset of our search space they
can express; and autoresearch~\citep{karpathy2025autoresearch} given an equal
wall-clock budget, which would measure how much the epoch-level granularity
actually buys. We report the multi-agent design as an architectural choice
supported by the per-axis ablation in Table~\ref{tab:ablation_study}, not as a
component proven necessary.

\textbf{Asymmetric search spaces.} LGT edits four axes jointly while several
baselines are structurally confined to one: DARTS to architecture, Optuna and
BOHB to hyperparameters. Part of our margin therefore reflects a larger action
space rather than better reasoning within it, and the two are not separated by
our current experiments.

\textbf{Scale.} Our benchmarks are moderate in size and the backbones compact.
Whether epoch-level editing remains stable when a single epoch costs hours, and
whether the Advisor's priors transfer to modern architectures at scale, is
untested; MLE-bench is the natural next target.

\textbf{Unbounded history.} The Advisor and Evaluator receive $\mathcal{H}_t$ in
full. Over long runs this grows past the context window, and we currently
truncate to the most recent entries rather than compacting or summarizing them.
A principled memory mechanism is future work.

\textbf{Constrained pipelines.} LGT assumes the configuration is genuinely free
to change. In settings built around fixed pretrained families and standardized
training stacks, most of our four axes are frozen by convention, and the method
reduces to hyperparameter tuning with a more expensive proposer.

\textbf{Fluent rationales are not verified rationales.} The Appendix~G examples
include a case where the Strategy Agent misstated the model's own dropout rate
while giving otherwise sensible advice. The validation gate blocks such errors
from affecting the result, but it does not block them from affecting a reader,
and the interpretability claim should be read with that in mind.

\end{document}